\documentclass{MITcsail}
\usepackage{microtype}
\usepackage{hyperref}
\usepackage{url}
\usepackage{booktabs}
\usepackage{xspace}
\usepackage{graphicx}
\usepackage{subcaption}
\usepackage{booktabs} 
\usepackage{forest}
\usepackage[inline]{enumitem}
\usepackage[most]{tcolorbox}
\usepackage{enumitem}
\usetikzlibrary{arrows.meta, shadows.blur}
\usepackage[frozencache=true,cachedir=minted-cache]{minted}
\usepackage{longtable}
\usepackage[capitalize,noabbrev]{cleveref}
\usepackage{tabularx} 
\usepackage{wrapfig}
\usepackage{needspace}

\makeatletter
\AddToHook{cmd/appendix/before}{\def\cref@section@alias{appendix}\def\cref@subsection@alias{appendix}}
\makeatother

\crefname{figure}{Fig.}{Figs.}
\Crefname{figure}{Fig.}{Figs.}
\crefname{section}{Sec.}{Secs.}
\Crefname{section}{Sec.}{Secs.}
\crefname{appendix}{App.}{Apps.}
\Crefname{appendix}{App.}{Apps.}

\newcommand{\tree}{thought-tree\xspace}
\newcommand{\trees}{thought-trees\xspace}

\makeatletter
\AddToHook{cmd/appendix/before}{\def\cref@section@alias{appendix}\def\cref@subsection@alias{appendix}}
\makeatother

\usepackage{lineno}

\definecolor{darkblue}{rgb}{0, 0, 0.5}
\hypersetup{colorlinks=true, citecolor=darkblue, linkcolor=darkblue, urlcolor=darkblue}

\title{Playing Psychic: Using Thought Trees to Predict Reasoning Models Accuracy on Coding Tasks}

\author
{Jiaxin Fang$^{1}$, Runyuan He$^{1}$, Sahil Bhatia$^{1}$, Neel Gajare$^{1}$, Alvin Cheung$^{1}$ \\
\vspace{1em} 
\normalfont{\small $^{1}$University of California, Berkeley}\\
}

\begin{document}

\maketitle
\thispagestyle{firstpagestyle} 

\begin{abstract}
Recent advances in large language models (LLMs) have shown that test-time scaling can substantially improve model performance on complex tasks, particularly in the coding domain. Under this paradigm, models use a larger token budget during inference to generate intermediate reasoning traces before producing a final answer. However, current evaluations primarily rely on competitive programming benchmarks, which may not capture the full range of reasoning abilities. In this work, we perform a systematic study of frontier reasoning models to understand their performance on real-world coding benchmarks. To gain more insights into the performance of such models, we devise a programmatic way to {\em automatically generate} coding tasks of arbitrary difficulty and structure from existing benchmarks. Using this framework, our analysis reveals that the structure of a reasoning trace, not just its contents, is a strong predictor of correctness. Motivated by this, we propose structured thought-trees as means to represent reasoning traces. To illustrate their use, we train a lightweight classifier on features extracted from thought-trees to predict trace correctness, and demonstrate that flagging and retrying structurally anomalous traces based on the extracted features yields consistent gains at lower complexity levels.
\end{abstract}

\section{Introduction}


Recent advances in large language models (LLMs) have shown that test-time scaling~\citep{guo2025deepseek, openai2024o1}, i.e., allocating larger inference budgets to generate intermediate reasoning before generating a final answer, can substantially improve performance on complex tasks, particularly in coding and mathematics. Under this paradigm, LLMs produce explicit reasoning traces, often in the form of chain-of-thought~\citep{wei2022cot} style decompositions before committing to a solution. This has led to major gains on benchmarks where solutions are verifiable, such as coding and mathematical problem solving.

Despite these advances, our understanding of the reasoning processes remains limited. Most evaluations focus on final answer correctness over a small set of domains, and provide little or no insights into the structure, quality or failure modes of the generated reasoning traces. Existing attempts to analyze reasoning traces~\citep{shojaee2025illusion, devic2025tracelength, zhao2025verifycot, aggarwal2025optimalthinking} have provided initial insights but remain limited in scope and methodology. These works largely rely on coarse heuristics such as trace length, token counts, self-consistency scores, or verifier-based correctness checks, and are often confined to narrow task domains such as competitive mathematics, symbolic reasoning, or single-task programming benchmarks. As a result, they offer limited understanding of the structure of reasoning traces, their generalization behavior across task distributions, and the systematic failure modes that emerge as reasoning complexity increases.

In this paper, we analyze reasoning traces through a structured and programmatic lens. Rather than the typical ``natural language to code'' task, we evaluate the frontier reasoning models on programming tasks that require {\em semantic reasoning} over programs, including code execution, fill-in-the-middle, and code translation.  A key challenge in LLM evaluation is finding benchmarks that are not already present in the training data. To address this, we introduce {\em programmatic task generation} that automatically creates new programming problems by intelligently composing existing ones. Doing so allows us to control the level of complexity as desired, and thus enables us to systematically evaluate different models across problem difficulty levels.

Another challenge in evaluating these traces at scale is that traces generated under test-time scaling are long and complex. Moreover, reasoning is inherently hierarchical and non-linear: LLMs explore alternatives, refine partial ideas, and backtrack. This makes the flat output token sequences a poor abstraction and tree-structured representation a natural abstraction for reasoning \citep{yao2023tree}. Motivated by this, we represent each reasoning trace as a {\em \tree}, where nodes correspond to atomic reasoning segments and edges encode semantic relationship between them. We construct these trees automatically using off-the-shelf LLMs, which incrementally organizes segments into a hierarchical structure. From these trees, we extract structural and semantic features that summarize how reasoning traces are organized. 

To demonstrate the usefulness of \trees, we extract structural and semantic features from \trees, and use such features to train lightweight classifiers to predict if a given reasoning trace correctly solves the programming task.
Surprisingly, our evaluation shows the extracted structural and semantic features are highly predictive of correctness, despite the simplicity of our classifiers' architecture. Across 3 different real-world programming benchmarks, our trained classifiers can correctly predict up to 89\% of them. We furthermore use the extracted features to augment the original prompt, and results show that the augmented prompts can improve the accuracy of the generated answers by up to 80\%.
Thought-trees thus act as structural probes and demonstrate that reasoning organization alone contains strong signal about success and failure of a trace.

In sum, we make the following contributions in this paper:
\begin{itemize}[leftmargin=*, itemsep=2pt]
    \item We propose programmatic benchmark generation (\cref{sec:prob_gen}) for coding tasks, to enable scalable and controllable evaluation beyond static datasets.
    \item We define structured \tree representation~(\cref{sec:tree_gen}) as a means to model reasoning traces, and we argue that \trees can be used to provide a principled abstraction for reasoning analysis.
    \item We leverage \tree to train structure-based classifiers (\cref{sec:classifier}). Results show that trace structure is a strong predictor of solution correctness. We show that classifier-identified structural failure signals can guide targeted intervention: retrying flagged traces yields consistent  gains at lower complexity levels 
\footnote{Our code and experiment results are available at: \href{https://github.com/lmcode-ai/thought-tree-analysis}{https://github.com/lmcode-ai/thought-tree-analysis}}
\end{itemize}
\section{Problem Formulation}
We consider an LLM evaluated under a test-time scaling paradigm, where the model generates an explicit reasoning trace prior to outputting the final solution.\\
Let $\mathcal{P} = \{p_1, \dots, p_N\}$
denote a set of coding problems, where each problem $p \in \mathcal{P}$ consists of a natural language specification and an associated correctness oracle.\\
Given a problem $p$, the LLM generates a \emph{reasoning trace} $\tau$
followed by a final program output $o(\tau;p)$. The trace may include intermediate reasoning steps, partial code, or planning tokens, and is generated using an expanded inference budget.\\
We define a binary correctness label:
$y(p, \tau) \in \{0,1\}$,
where $y=1$ if the output $o(\tau;p)$ solves problem $p$ according to the oracle, and $y=0$ otherwise. \\
Instead of treating $\tau$ as a flat token sequence, we represent each reasoning trace using a structured abstraction, referred to as a \emph{thought-tree}.  Formally, a thought-tree is a rooted tree $T(\tau) = (V,E)$, where each node in $V$ corresponds to reasoning segments and each directed edge $(u,v) \in E$ means that node $v$ is a refinement or continuation of node $u$. \\
For each reasoning trace $\tau$, we extract a feature vector
%
$\phi(\tau) = \big[\phi_{\text{tree}}(T(\tau)) \;\Vert\; \phi_{\text{trace}}(\tau)\big] \in \mathbb{R}^d$,
where $\phi_{\text{tree}}$ captures structural properties of the thought-tree, and $\phi_{\text{trace}}$ captures features derived directly from the raw reasoning trace. Here $\Vert$ denotes feature concatenation.\\
Given a dataset of problems and generated traces
$
\mathcal{D} = \{(p_i, \tau_i, y_i)\}_{i=1}^M
$,
our goal is to learn a lightweight classifier
$
f_\theta : \mathbb{R}^d \rightarrow \{0,1\}
$
that predicts whether a reasoning trace leads to a correct solution:
$
f_\theta(\phi(\tau)) \approx y(p, \tau)
$.
The classifier parameters $\theta$ are learned by minimizing a standard empirical risk objective,
$
\min_\theta \; \frac{1}{M} \sum_{i=1}^M \ell\big(f_\theta(\phi(\tau_i)), y_i\big)
$,
where $\ell(\cdot,\cdot)$ is a binary classification loss.

\begin{figure*}
  \centering

  \textbf{Solve f1(f2(text))}\\
  \vspace{0.5em}
  \begin{minipage}[b]{0.48\textwidth}
    \begin{minted}[fontsize=\scriptsize,breaklines,frame=lines]{python}
    def f1(text: str) -> str:
      i = 0
      while i < len(text) and text[i].isspace():
        i+=1
      if i == len(text):        
        return 'space'   
      return 'no'
    \end{minted}
    \subcaption{Source Code 1}
    \label{fig:src1}
  \end{minipage}
  \hfill
  \begin{minipage}[b]{0.48\textwidth}
    \begin{minted}[fontsize=\scriptsize,breaklines,frame=lines]{python}
    def f2(text: str) -> list[int]:
      occ = {}
      for ch in text:
        n = {'a': 'b', 'b': 'c', 'c': 'd', 'd': 'e'}
        n = n.get(ch, ch)
        occ[n] = occ.get(n, 0) + 1
      return [x for _, x in occ.items()]
    \end{minted}
    \subcaption{Source Code 2}
    \label{fig:src2}    
  \end{minipage}

  \textbf{Workflow Step: Segmentation}
  \vspace{0.5em}

  \fbox{
  \begin{minipage}{0.45\textwidth}
    \small
    \textbf{Raw Reasoning Trace}\\
    We are given two functions: f1 and f2.
    
    First, let's analyze f1:\\
      f1(text) does:\\
        - It skips any leading whitespace by incrementing i until a non-space character is found or the end of the string.\\
        - If the entire string is spaces (i.e., i reaches the length of the text), it returns the string 'space'.\\
        - Otherwise, it returns 'no'.\\  
    Now, we are to compute: f2(f1("     "))\\
  \dots
  \end{minipage}}
  \hfill
  $\Rightarrow$
  \hfill
  \fbox{
  \begin{minipage}{0.45\textwidth}
    \small
    \textbf{Segmentation Results}\\
    Seg 0: "We are given two functions: f1 and f2."\\
    Seg 1: "First, let's analyze f1: \textbackslash n  f1(text)" \\
    Seg 2: "does: \textbackslash n    - It skips any leading whitespace by incrementing i until a non-space character is found or the end of the string.\textbackslash n    - If the entire string is spaces (i.e., i reaches the length of the text), it returns the string 'space'.\textbackslash n    - Otherwise, it returns 'no'."\\
    Seg 3: "Now, we are to compute: f2(f1("     ")) Step 1: Evaluate f1("     ")"\\
    \dots
  \end{minipage}}
  \vspace{-0.1in}
  \caption{Overview of the approach: Input code is analyzed and segmented to produce atomic reasoning units.}
  \label{fig:workflow_top}
  \vspace{-0.2in}
\end{figure*}

\section{Approach}
\label{sec:approach}
We illustrate our overall approach in~\cref{fig:workflow_top} and ~\cref{fig:workflow_tree}. It consists of three stages: programmatic problem generation, converting raw reasoning traces to \tree and learning classifiers to predict which we describe next in detail.
\subsection{Programmatic Problem Generation}
\label{sec:prob_gen}
LLMs are primarily trained and evaluated on natural language (NL) $\rightarrow$ code tasks~\citep{hendrycks2021apps, jain2024livecodebench, chen2021codex, austin2021mbpp}. While strong performance on such benchmarks demonstrates the model's code generation capabilities, it provides a limited view of the model's reasoning abilities on a broader class of programming problems.\\
We focus on 3 canonical programming task families: code execution reasoning, code translation, and fill-in-the-middle (FIM) completion. These tasks differ from the classical NL $\rightarrow$ code tasks as they provide a structured programming interface, enabling systematic composition and the generation of new problems. We use them as atomic building blocks to construct structured, multi-stage coding tasks.\\
We construct new coding problems by composing existing programs through \emph{type inference}. Each base problem $p$ is first annotated with its input and output types, resulting in a typed signature:
$
f_p : (x_1: \tau_1, \dots, x_k: \tau_k) \rightarrow y: \tau_{out}
$
Given a problem $p_i$, we select another problem $p_j$ such that the output type of $p_i$ matches the input signature of  $p_j$, and define a composed program as:
$
f_{ij}(x) = f_{p_j}(f_{p_i}(x))
$
This can be applied recursively to form composition chains:
$
f_{1:n}(x)= f_{p_n}\circ \dotsb \circ f_{p_1}(x) 
$
yielding new problems whose solutions require solving multiple subprograms in sequence. \\
Our construction has two key advantages:
\begin{enumerate*}
    \item controllable composition depth, which enables systematic scaling of task complexity, and
    \item inherited correctness oracles, since composed tasks remain automatically evaluable through execution of their constituent programs.
\end{enumerate*}\\
\Cref{fig:src1,fig:src2} illustrate two example base programs from the code execution reasoning dataset. Function $f_1: str \rightarrow str$ maps an input string to a categorical output (\texttt{"space"} or \texttt{"no"}), while $f_2:str \rightarrow list[int]$ performs a character-level transformation and frequency aggregation. Since the output type of $f_1$ matches the input type of $f_2$, the programs are type-compatible and can be composed to form a higher-order task $f_2 \circ f_1$. The resulting task is: given an input to $f_1$, predict the output of $f_2(f_1(\cdot))$ which requires multi-stage reasoning over both subprograms.\\
Unlike manual problem curation or community-sourced benchmarks, this approach provides a programmatic, scalable, controllable, and automatically evaluable mechanism for generating structured reasoning tasks.

\subsection{Tree Generation}
\label{sec:tree_gen}

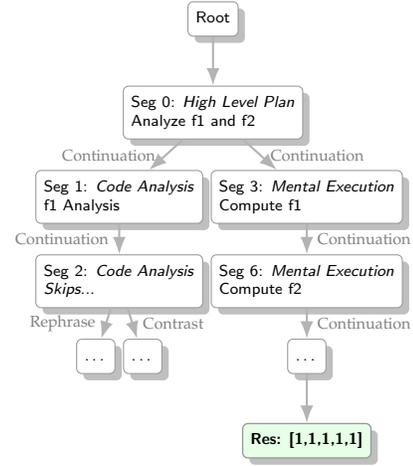
\begin{wrapfigure}{r}{0.45\textwidth}
    \centering
    \vspace{-10pt}
    \begin{forest}
      for tree={
        font=\sffamily\tiny,
        draw=gray!80,
        fill=white,
        rounded corners=2pt,
        edge={-Latex, thick, gray!60},
        align=left,
        inner sep=3pt,
        l sep=0.4cm,
        s sep=0.1cm,
        drop shadow
      }
      [Root
        [Seg 0: \textit{High Level Plan} \\ Analyze f1 and f2
          [Seg 1: \textit{Code Analysis} \\ f1 Analysis,  edge label={node[midway, left, text=gray,] {\tiny Continuation}}
            [Seg 2: \textit{Code Analysis} \\ \textit{Skips...}, edge label={node[midway, left, text=gray,] {\tiny Continuation}}
            [\dots, edge label={node[midway, left, text=gray,] {\tiny Rephrase}}]
            [\dots, edge label={node[midway, right, text=gray,] {\tiny Contrast}}] ]
          ]
          [Seg 3: \textit{Mental Execution} \\ Compute f1, edge label={node[midway, right, text=gray,] {\tiny Continuation}}
               [Seg 6: \textit{Mental Execution} \\ Compute f2, edge label={node[midway, right, text=gray,] {\tiny Continuation}}
                   [\dots, edge label={node[midway, right, text=gray] {\tiny Continuation}}
                     [\textbf{Res: [1,1,1,1,1]}, fill=green!10, line width=0.5pt]
                    ]  
               ]
          ]
        ]
      ]
    \end{forest}
    \caption{Hierarchical thought tree.}
    \label{fig:workflow_tree}
    \vspace{-0.2in}
\end{wrapfigure}

For each problem $p$, the model generates a reasoning trace $\tau$ under test-time scaling, which we transform into a structured \tree representation through three sequential steps: segmentation, semantic labeling, and tree construction (illustrated in \cref{fig:workflow_top,fig:workflow_tree}).

\textbf{Segmentation.} \Cref{fig:workflow_top} shows the segmentation process. On the left, we show a sample of the raw reasoning trace $\tau$ for the composed problem described in the previous section, while the right shows the resulting decomposition into atomic thought segments. Each segment corresponds to a coherent reasoning unit (e.g., setup, function analysis, logic description, or execution planning), demonstrating how a flat trace is transformed into structured, self-contained reasoning components. We decompose $\tau$ into a sequence of atomic thought units $S(\tau) = (s_1, s_2, \dots, s_K)$ using a rule-based approach: sentences are greedily merged based on transition signals (e.g., ``\textit{wait}'', ``\textit{alternatively}''), sentence length, and code-span boundaries.
\needspace{5\baselineskip}

\textbf{Semantic Labeling.}
\label{sec:sem_label}
Each segment $s_i$ is assigned a label $l(s_i) \in \mathcal{L}$ by an off-the-shelf LLM (prompt in \cref{app:prompts}), where $\mathcal{L} = \{$\textit{code analysis, mental execution, test generation, bug fixing, language mapping, high level plan, empty}$\}$ captures the functional role of each reasoning step. These labels are not used during tree construction; instead, they serve as distributional features for the classifier (\cref{sec:classifier}).

\textbf{Tree Construction.}
\Cref{fig:workflow_tree} illustrates an example of the resulting hierarchical \tree. Each node corresponds to a segmented reasoning unit, and the tree structure makes explicit how the model’s reasoning evolves through analysis, execution, and composition, transforming a flat trace into an interpretable hierarchical representation. Segments are organized into a rooted tree $T(\tau) = (V, E)$ by an off-the-shelf LLM that processes segments sequentially, attaching each to an existing node with a relation from $\mathcal{R} = \{\textit{Continuation, Contrast, Rephrase}\}$. These relations capture whether a segment extends, opposes, or rephrases its parent. The resulting tree makes explicit how reasoning evolves through analysis, execution, and backtracking, and serves as the basis for feature extraction.

\subsection{Analyzing Thought-Trees}
\label{sec:classifier}
We manually examine the \trees and design a number of feature groups that can be used to analyze them. Specifically, for each reasoning trace $\tau$, we extract a feature vector $\phi(\tau) \in \mathbb{R}^d$ consisting of three feature groups:

\noindent\textbf{Structural tree features:}
    \begin{itemize}[nosep,leftmargin=1em,labelwidth=*,align=left]
        \item Tree depth and size (e.g., maximum depth, total number of nodes), reflecting overall reasoning depth and structural complexity.
        \item Branching behavior (e.g., average branching factor, root branching factor), capturing exploration and early-stage divergence.
        \item Depth-normalized statistics (e.g., node distributions per depth, internal nodes per depth), describing how reasoning effort is allocated across stages.
        \item Internal structure metrics (e.g., \# of internal/non-leaf nodes), capturing the extent of intermediate reasoning.
    \end{itemize}

\noindent\textbf{Raw Trace-level features:}
    \begin{itemize}[nosep,leftmargin=1em,labelwidth=*,align=left]
        \item Exploration and revision signals, such as explicit backtracking events and forward-progress markers, capturing whether the model revises earlier decisions or commits to a reasoning path.
        \item Lexical diversity and confidence-related features (e.g., unique word ratio, success word count), capturing variation and certainty in expression. 
        \item Length and density statistics (e.g., token counts, code-to-text ratio), characterizing trace verbosity and structure.
    \end{itemize}

\noindent\textbf{Semantic label features:} We add the semantic labels (described in~\cref{sec:sem_label}) assigned to each segment as distributional features. These features provide additional signals beyond purely structural properties. \\
The final feature vector is:
$\phi(\tau) = [\phi_{tree}(T(\tau)) \;\Vert\; \phi_{trace}(\tau) \;\Vert\; \phi_{label}(\tau)]
$.

To study \trees, we use the extracted features for 2 purposes: first, we train a lightweight classifier $f_{\theta}$ on $\phi(\tau)$ to predict trace correctness. The model is intentionally simple, as our goal is not to maximize predictive capacity, but to evaluate if structural properties of reasoning contain sufficient signal to predict correctness. Next, we use $\phi(\tau)$ for intervention by augmenting the prompts to see if they can improve accuracy (\cref{sec:intervention}).

\section{Experiments}

We evaluate on three coding benchmarks: CRUXEval~\citep{gu2024cruxeval}, SAFIM~\citep{gong2024safim}, and CodeLingua~\citep{pan2024codelingua}. Details of the benchmarks are provided in \cref{sec: dataset}.

\begin{figure*}[h] 
\centering
\footnotesize
\begin{tabularx}{\textwidth}{l l l X} 
\toprule
\textbf{Benchmark} & \textbf{Task Type} & \textbf{Size} & \textbf{Description} \\
\midrule
\textbf{CRUXEval} & Code execution reasoning & 800 & Predict output of a Python function on a given input \\ \addlinespace
\textbf{SAFIM} & Fill-in-the-middle & 17,720 & Generate missing code blocks from context \\ \addlinespace
\textbf{CodeLingua} & Code translation & 1,700 & Translate code across languages preserving semantics \\
\bottomrule
\end{tabularx}
\caption{Benchmarks used in our evaluation.}
\label{tab:benchmarks}
\end{figure*}

\begin{figure*}[t]
\centering
\footnotesize
\begin{tabular}{ll ccc ccc}
\toprule
& & \multicolumn{3}{c}{\textbf{DeepSeek-R1}} & \multicolumn{3}{c}{\textbf{QwQ-32B}} \\
\cmidrule(r){3-5} \cmidrule(l){6-8}
\textbf{Benchmark} & \textbf{Task} & \textbf{L1} & \textbf{L2} & \textbf{L3} & \textbf{L1} & \textbf{L2} & \textbf{L3} \\
\midrule
CRUXEval & Output & 94.13 & 79.13 & 79.25 & 82.63 & 74.50 & 72.38 \\
\midrule
SAFIM & Block & 69.47 & 17.38 & 16.24 & 48.90 & 12.92 & 9.35 \\
\midrule
CodeLingua & Java $\rightarrow$ C & 94.00 & 91.60 & 90.00 & 84.44 & 72.00 & 55.02 \\
& Python $\rightarrow$ C & 82.40 & 86.00 & 85.20 & 65.60 & 60.24 & 46.00 \\
& Java $\rightarrow$ Fortran & 58.40 & 52.80 & 48.00 & 42.40 & 27.60 & 11.29 \\
\bottomrule
\end{tabular}
\vspace{-0.1in}
\caption{Pass@1 (\%) results across benchmarks and composition levels.}
\label{tab:pass1_results}
\vspace{-0.25in}
\end{figure*}

\subsection{Experiment Setup} 
\label{sec:exp_setup}

\textbf{Problem Generation.}
Each problem in the three selected benchmarks is treated as a Level-1 (L1) task. Following the approach detailed in~\cref{sec:prob_gen}, we generate higher-complexity datasets by function composition: Level-2 (L2) tasks are formed by composing two base programs, and Level-3 (L3) tasks by composing three base programs.

\textbf{Evaluating LLMs.}
For every problem across all composition levels (L1, L2, and L3), we evaluate DeepSeek-R1~\citep{guo2025deepseek} and QwQ-32B~\citep{yang2025qwen3} using their default temperature settings under a test-time scaling regime. Prompt templates used for evaluating the models are provided in~\cref{app:prompts}.

\textbf{Tree Generation.}
The collected reasoning traces are converted to \tree representations using the approach described in~\cref{sec:tree_gen}. Segmentation, semantic labeling, and tree construction are done using Gemini 2.0 Flash and Qwen3-235B-A22B-Instruct (temperature = 0.2). From each trace, we extract a feature set consisting of tree-structural, trace-level and semantic label features, as described in~\cref{sec:classifier}.

\textbf{Classifier Learning.}
\label{sec: classifier_setup}
We train lightweight Random Forest classifiers in Python \texttt{scikit-learn} library to predict whether a given reasoning trace produces a correct program output. Classification is formulated as a supervised binary prediction task over feature vectors derived from \tree and traces.  We adopt a cross-level generalization setup: classifiers are trained exclusively on 80\% of the \textbf{L1} traces and evaluated the trained classifiers on the remaining 20\% of \textbf{L1}, \textbf{L2}, and \textbf{L3} traces. For each benchmark, we use Random Forest classifiers with 50 estimators, and tune the maximum tree depth.

\begin{figure}[h]
\vspace{-0.1in}
    \begin{tcolorbox}[title=Model Output Trace] 
    \small 
        \textbf{[Step-by-Step Stagnation]} \dots Wait, the elements are: Indices 0:9, 1:7 \dots Wait let me do it step by step \dots Wait let's count each element in reverse \dots \\
        \textbf{[Hypothesis Fixation]} Wait, but the loop is up to len(digits) \dots  \\
        \textbf{[Infinite Loop Phase]} \textit{Hmm. Alternatively, perhaps the code is correct and the error is not there. Hmm. Alternatively, perhaps the code is correct and the error is not there. [Repeated till the end] \dots} 
    \end{tcolorbox}
    \vspace{-0.1in}
    \caption{Example of a terminal reasoning loop where brute-force manual execution triggers a collapse into infinite repetition.}
    \label{exp: loop}
    \vspace{-0.2in}
\end{figure}

\subsection{Evaluation Results}
\label{sec: eval}
We first evaluate the reasoning models on the composed benchmarks to evaluate their robustness under increasing compositional complexity. In \cref{tab:pass1_results}, the performance degradation follows a distinct "cliff" pattern that varies by benchmark complexity. For CRUXEval, the drop-off is relatively small; given good performance on L1 baseline and the simplicity of short Python functions, both R1 and QwQ successfully solve chained functions one-by-one with minimal state-tracking overhead.\\
In contrast, SAFIM exhibits a near-vertical decline. This suggests that for "block" reasoning—which requires infilling code based on long-range dependencies—the compositional overhead becomes multiplicative. Manual inspection reveals two primary failure modes in higher-level SAFIM tasks: (1) State-Tracking Exhaustion: the model fails to correctly propagate input/ouput between functions, and (2) Logical Recursion: the model often enters repetitive reasoning loops when attempting to track nested dependencies, leading to token limit exhaustion before a conclusion is reached (\cref{exp: loop}) . These results indicate that while LLMs excel at local function logic, their ``working memory'' for nested functional states remains a primary bottleneck. \\
The CodeLingua results reveal a divergence in compositional robustness between the two models. In high-resource pairs like Java $\rightarrow$ C and Python $\rightarrow$ C, R1 remains stable, preserving semantic invariance across nested layers. Conversely, QwQ’s accuracy decays sharply, likely due to its much smaller parameter scale leading to multiplicative error accumulation. This ``compositional breaking point" is further evidenced in Java $\rightarrow$ Fortran, where even R1’s performance drops significantly. In low-resource contexts like Fortran, the lack of foundational proficiency acts as a bottleneck; the additional cognitive overhead of functional composition exceeds the model's reasoning capacity, leading to a performance collapse.

\begin{figure*}[t]
\centering
\small
    \centering
    \begin{tabular}{llcccc}
        \toprule
        \textbf{Dataset} & \textbf{Level} & \textbf{R1 Acc.} & \textbf{R1 Interv. Pass@1} & \textbf{QwQ Acc.} & \textbf{QwQ Interv. Pass@1}\\ \midrule
        CRUXEval & L1 & 0.88 & 0.80 (16/20) & 0.89 & 0.65 (42/65) \\
                 & L2 & 0.75 & 0.22 (11/51) & 0.77 & 0.13 (4/30)\\
                 & L3 & 0.73 & 0.15 (10/66) & 0.74 & 0.27 (28/104)\\ \midrule
        SAFIM    & L1 & 0.79 & 0.39 (47/119) & 0.79 & 0.16 (39/241) \\
                 & L2 & 0.83 & 0.21 (22/106) & 0.74 & 0.06 (25/397) \\
                 & L3 & 0.81 & 0.17 (13/78) & 0.71 & 0.04 (13/332) \\ \midrule
        CodeLingua & L1 & 0.73 & - & 0.68 & -\\
                 & L2 & 0.75 & - & 0.68 & -\\
                 & L3 & 0.76 & - & 0.65 & -\\ \bottomrule
    \end{tabular}
\vspace{-0.1in}
\caption{\small Random Forest classification results and intervention performance across three composition levels. Classifiers were independently trained and validated on Level 1 (L1) data for R1 and QwQ, with the resulting classifiers evaluated on Level 2 (L2) and Level 3 (L3) data to assess generalization. We further report the Intervention Pass@1, representing the success rate of the model in recovering the correct output after the classifier identifies a potential failure.}
\label{tab: classi_interv}
\end{figure*}

\subsection{Reasoning Trace Analysis}
\subsubsection{Classification and Feature Analysis.}
To understand whether observed performance ``cliffs" are reflected in models' internal logic, we trained Random Forest classifiers to predict output correctness from reasoning trace features (\cref{sec: classifier_setup}).
As shown in \cref{tab: classi_interv}, classifiers generalize well across complexity levels, though with benchmark-specific trends. For \textbf{CRUXEval}, accuracy is strongest at L1 and degrades gradually at L2--L3 as task complexity increases, yet remains relatively high throughout. For \textbf{SAFIM}, its balanced correct/incorrect trace distribution enables the classifier to learn diverse failure cases effectively, with accuracy remaining stable across all levels. For \textbf{CodeLingua}, R1 accuracy improves slightly from L1 to L3, suggesting that correctness signals learned at L1 are transferable and become more distinct at higher complexity; QwQ exhibits a mild downward trajectory toward L3, indicating that correct-trace markers become harder to identify as complexity grows.\\
We use Partial Dependence Plots (PDPs) to analyze the marginal effect of each feature on predicted failure probability, visualizing the top-six features per classifier (\cref{fig:pdp_comparison}).

\label{sec:features}
\textbf{Common Features.} Structural tree features, such as total segments, internal and total nodes, are positively associated with error probability, suggesting that deeper, more fragmented reasoning trees increase failure risk. Length-based metrics (log word count, log length) similarly correlate with higher failure rates, while greater lexical diversity is associated with lower error probability, indicating that vocabulary-rich, high-level reasoning is a marker of robustness.

\textbf{CRUXEval.} Mental execution (brute-force simulation steps) is positively correlated with error probability and strongly co-varies with word count (\cref{fig:mental_exec_word_count}), consistent with ``overthinking'' on straightforward tasks. Conversely, a higher analysis rate (statements referencing the code directly) is negatively correlated with failure, indicating that grounded code analysis, rather than exhaustive simulation, drives correctness.

\textbf{CodeLingua.} Frequent code-analysis segments are strongly associated with failure for both R1 and QwQ, suggesting the model stalls in repeated source code description rather than progressing toward actual translation. For QwQ, excessive high-level planning similarly predicts failure, while higher code density, which reflects concrete implementation logic, is associated with correctness. Structurally, depth is preferable to breadth: deep trees reflect coherent reasoning, whereas excessive branching produces fragmented traces that impede translation.

\subsubsection{Intervention}
\label{sec:intervention}
To assess the practical utility of the features, we augment the prompt (shown in~\cref{sec:intervention_prompt}) to retry traces that are truly incorrect and statistically distinguishable as failures, as characterized by the predictive features in \cref{sec:features}. Specifically, a trace is flagged for retry if it exhibits a z-score greater than 1.96 on at least one common predictive feature, indicating statistically significant deviation from the distribution of correct traces. We do not run intervention on CodeLingua, as the proportion of flagged traces meeting this criterion is too small to yield meaningful recovery statistics. We report the recovery rate as Intervention Pass@1 in \cref{tab: classi_interv}. Across CRUXEval and SAFIM, intervention is most effective at L1, where the model successfully recovers correct outputs at a high rate. However, recovery degrades sharply at higher composition levels, suggesting that failures at L2 and L3 reflect deeper reasoning breakdowns that a simple retry cannot resolve. This decay mirrors the performance cliff observed in \cref{tab:pass1_results}, reinforcing that compositional complexity introduces systematic failure modes rather than incidental errors.

\subsection{Ablation Studies}

\subsubsection{An Oracle Classifier}
\label{sec:oracle_classifier}
To evaluate the robustness and upper-bound performance of the reasoning trace classifiers, an ablation study was conducted comparing a level-specific transfer approach against an ``Oracle" baseline. In the absence of a ground-truth standard, the Oracle classifier represents a theoretical ceiling; it is trained and validated on a pooled dataset containing all composition levels (L1, L2 and L3), ensuring exposure to the full distribution of task complexities (\cref{app:oracle}). This benchmark is then compared to the results in \cref{tab: classi_interv}.\\
We find that training on L2 and L3 data does not yield substantial performance gains: accuracy remains largely consistent across CRUXEval, SAFIM, and CodeLingua. This indicates that single-level (L1) training already captures the core indicators of reasoning success and failure, enabling effective prediction even in highly composed tasks without level-specific training data.

\begin{figure*}[t]
\centering
\begin{tikzpicture}
\node[anchor=south, inner sep=4pt] (legend) at (0.5\linewidth, 0) {
  \footnotesize
  \begin{tabular}{@{}ccccc@{}}
    \tikz\draw[blue!70, solid, line width=1.5pt] (0,0) -- (0.5,0); \textcolor{blue!70}{L1} &
    \tikz\draw[orange!80, solid, line width=1.5pt] (0,0) -- (0.5,0); \textcolor{orange!80}{L2} &
    \tikz\draw[green!60!black, solid, line width=1.5pt] (0,0) -- (0.5,0); \textcolor{green!60!black}{L3} &
    \tikz\draw[gray, dotted, line width=1.5pt] (0,0) -- (0.5,0); Length Only &
    \tikz\draw[gray, dashed, line width=1.5pt] (0,0) -- (0.5,0); Non-Tree
  \end{tabular}
};
\node[anchor=north, inner sep=0] at ([yshift=-4pt]legend.south) {
  \includegraphics[width=\linewidth]{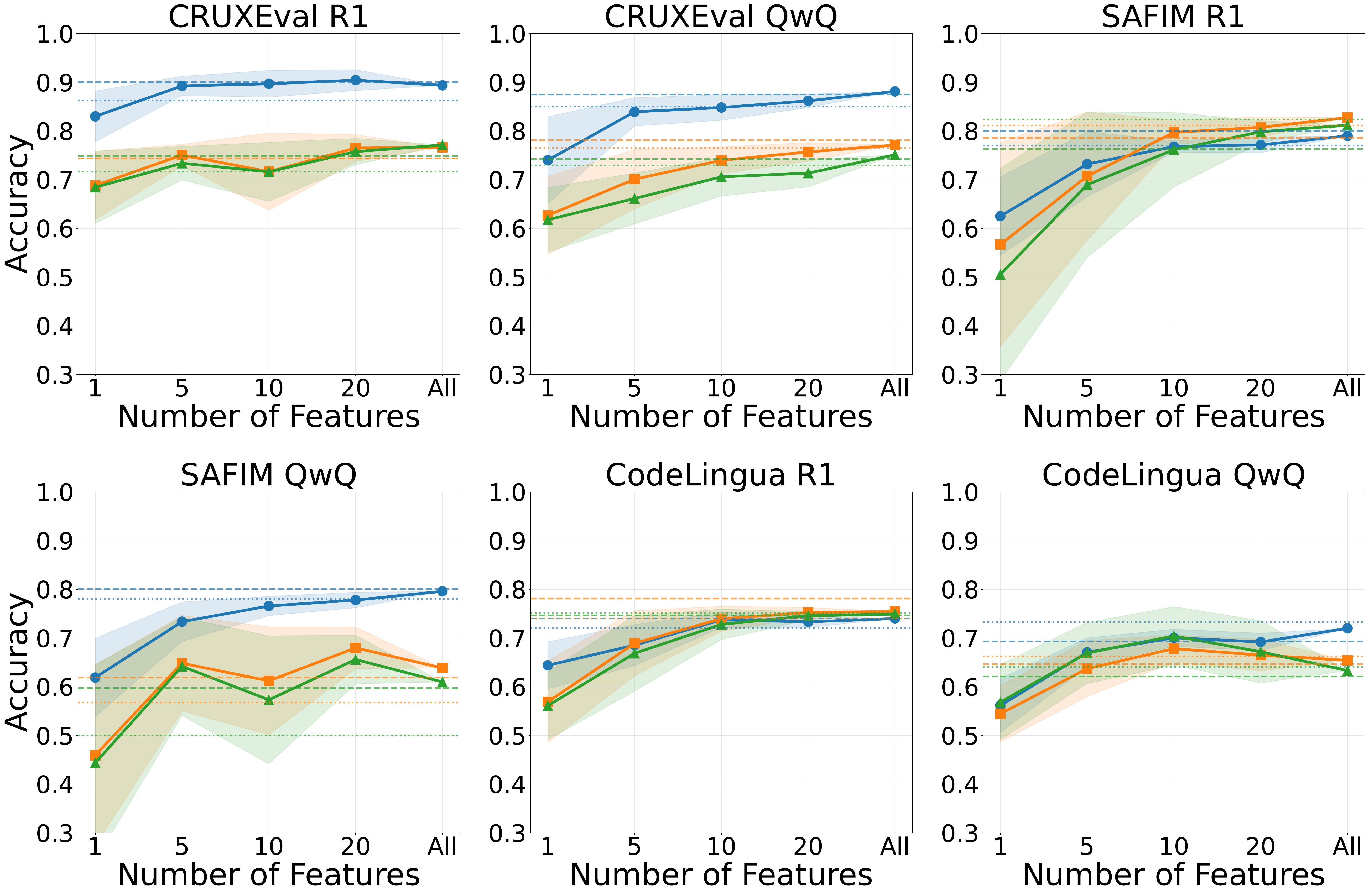}
};
\end{tikzpicture}
\caption{\small Classification accuracy as a function of randomly sampled feature count. Shaded regions indicate $\pm$1 standard deviation across 10 random sampling trials. Dotted and dashed horizontal lines show the performance of Length Only (9 features) and Non-Tree (44 features) subsets, respectively.}
\label{fig:feature_ablation}
\end{figure*}
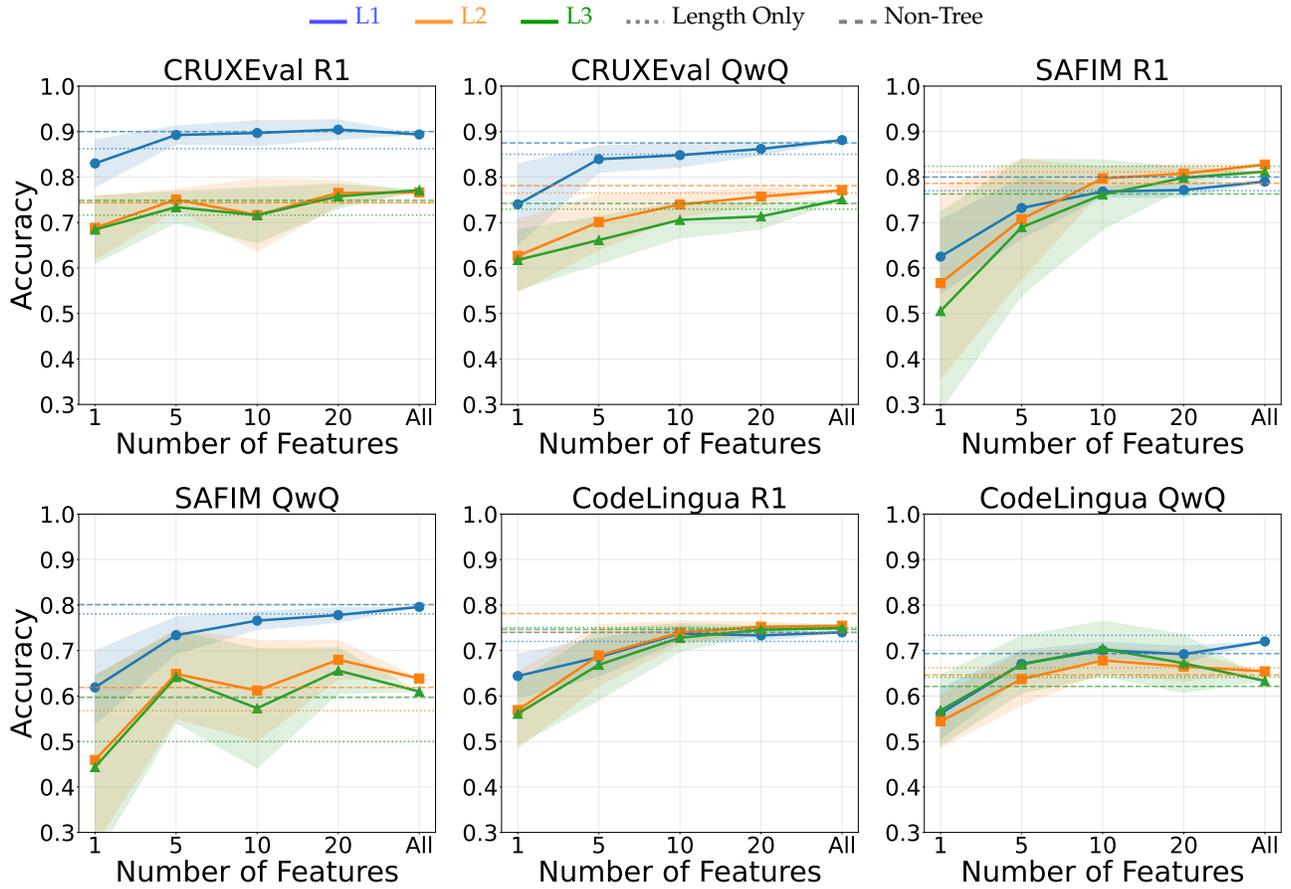
\subsubsection{Random Feature Sampling}

To further investigate feature redundancy and the marginal contribution of additional features, we conduct a random sampling experiment across all six dataset-model combinations. Specifically, we randomly sample 1, 5, 10, 20, and all features, repeating each configuration 10 times to obtain mean and standard deviation estimates. We additionally evaluate two fixed feature subsets: \emph{Length Only} (9 length-related features such as \texttt{log\_length}, \texttt{log\_word\_count}, and segment/sentence length statistics) and \emph{Non-Tree} (all features except the 10 tree-structure features). The results are shown in \cref{fig:feature_ablation}.

Several findings emerge from this analysis. First, the high variance observed with single-feature sampling (standard deviation ranging from 0.05 to 0.22) confirms that feature importance varies substantially---certain features carry significantly more predictive power than others. Second, classification performance generally saturates at 10--20 features, suggesting that a small subset of structural features captures most of the discriminative information.

However, the fixed feature subset results reveal that neither length-based nor non-tree features alone provide consistently reliable performance. The Length Only subset exhibits dramatic accuracy drops across multiple configurations: on SAFIM QwQ, accuracy at L2 and L3 falls well below the full model, and similar regressions are observed on CRUXEval QwQ at higher composition levels, where the gap between Length Only and the full model widens substantially as task complexity increases. The Non-Tree subset similarly regresses on SAFIM R1, demonstrating that discarding tree-structure information introduces meaningful performance losses even when other features are retained. These regressions illustrate that relying on any single feature category introduces blind spots: length features miss structural signals critical for cross-level transfer, while removing tree features discards information essential for maintaining generalization accuracy. Consequently, the complete feature set remains necessary to avoid these performance regressions across diverse dataset-model-level configurations.
\section{Related Work}
\textbf{Reasoning Models.}
Recent work has focused on building LLMs that explicitly generate intermediate reasoning before generating the final answer. Models such as OpenAI's o1 family~\citep{openai2024o1} and DeepSeek-R1~\citep{guo2025deepseek} use reinforcement learning and test-time scaling to induce emergent reasoning behaviors, achieving strong performance on verifiable domains such as math and coding. However, these efforts primarily focus on improving task performance rather than analyzing the structure, organization and failure modes of the reasoning traces themselves.

\textbf{Analyzing Reasoning Traces.}
\citet{shojaee2025illusion} study reasoning under increasing complexity but focus on performance collapse rather than structural representation. \citet{devic2025tracelength} use trace length as an uncertainty signal, relying on coarse heuristics. \citet{zhao2025verifycot} apply graph-based verification to assess reasoning correctness, treating structure as a verification tool rather than a representation for learning reasoning organization. Verification targets internal logical validity, whereas our work uses reasoning structure as a representational signal for learning and prediction. \citet{aggarwal2025optimalthinking} analyze under-/over-thinking using task-specific heuristics without a general structural abstraction. In contrast, we introduce a structured thought-tree representation and show that the organization of reasoning itself predicts correctness across tasks of controlled compositional complexity.

\section{Conclusion}
We present a principled study of frontier reasoning models on coding tasks. To this end, we design an algorithm that automatically generates new coding programs from existing ones. To systematically analyze reasoning traces, we propose \emph{\trees} as a representation for each trace. Training a simple model on \trees demonstrates their effectiveness for such analysis and sheds light on the behavior of reasoning models in coding tasks.


\bibliography{references}
\bibliographystyle{plainnat}

\appendix
\newpage
\appendix


\section{Dataset Description}
\label{sec: dataset}
The datasets are selected to probe "inner-logic" capabilities. They provide a holistic view of whether a model truly understands the mechanics of code or is simply relying on high-probability pattern matching from its training data.
\begin{enumerate}[nosep,leftmargin=1.5em,labelwidth=*,align=left]
    \item \textbf{CRUXEval}~\citep{gu2024cruxeval} is a code execution reasoning benchmark that consists of 800 standalone Python functions. The task is to predict the output of a given function on a specific input.
    \item \textbf{SAFIM}~\citep{gong2024safim} is a fill-in-the-middle benchmark and includes 17,720 examples from Java, C, C\# and Python with tasks designed for the syntax-aware completion of program structures. In this paper, we evaluate the block completion, a task requiring the model to generate an entire code block based on surrounding context and problem background.
    \item \textbf{Codelingua}~\citep{pan2024codelingua}  involves the translation of 1,700 code samples from three benchmarks and two real-world projects. The task asks a model to read a snippet in a source language and reconstruct it in a target language without altering the underlying logic and final outputs. A number of different programming languages are represented in the source and target.
\end{enumerate}

\subsection{CRUXEval Example Problem}
\begin{tcolorbox}
\begin{verbatim}
def f(nums):
    output = []
    for n in nums:
        output.append((nums.count(n), n))
    output.sort(reverse=True)
    return output
\end{verbatim}
Input: [1, 1, 3, 1, 3, 1]\\
Output: [(4, 1), (4, 1), (4, 1), (4, 1), (2, 3), (2, 3)]
\end{tcolorbox}

\subsection{Codelingua Example Problem}
\begin{tcolorbox}
Source Language: Python\\
Target Language: C
        \begin{verbatim}
def f(nums): 
    output = []    
    for n in nums:
        output.append((nums.count(n), n))
        output.sort(reverse=True)    
    return output
        \end{verbatim}
\end{tcolorbox}

\subsection{SAFIM Example Problem}
\begin{tcolorbox}
Fill in the blank code.\\
Description: 
        You are given three integers $a$, $b$, and $c$.
Determine if one of them is the sum of the other two.

Input Specification:
The first line contains an integer $t$ ($1 \leq t \leq 9261$),
the number of test cases. Each test case contains three integers
$a$, $b$, and $c$ ($0 \leq a,b,c \leq 20$).

Output Specification:
For each test case, output \texttt{YES} if one number is the sum of
the other two, and \texttt{NO} otherwise.
\begin{verbatim}
import java.util.*;

public class A2 {
    public static void main(String[] args) {
        Scanner in = new Scanner(System.in);
        for (int t = in.nextInt(); t > 0; t--) {
            int a = in.nextInt();
            int b = in.nextInt();
            int c = in.nextInt();

            if (a + b == c || a + c == b || b + c == a) {
                System.out.println("YES");
            } else {
                System.out.println("NO");
            }
        }
    }
}
\end{verbatim}
Ground Truth: 
\begin{verbatim}
int n1 = in.nextInt(), n2 = in.nextInt(), n3 = in.nextInt();
double sum = n1+n2+n3;
System.out.println((sum/2 == n1 || sum/2 == n2 || sum/2 == n3) ? "YES" : "NO"); 
\end{verbatim}
Unit Tests: [{"input": "7 1 4 3 2 5 8 9 11 20 0 0 0 20 20 20 4 12 3 15 7 8", "output": ["YES NO YES YES NO NO YES"]}]
\end{tcolorbox}

\newpage
\section{Partial Dependence Plot}
\begin{figure*}[h]
    \centering
    \begin{subfigure}[b]{0.32\textwidth}
        \centering
        \includegraphics[width=\linewidth]{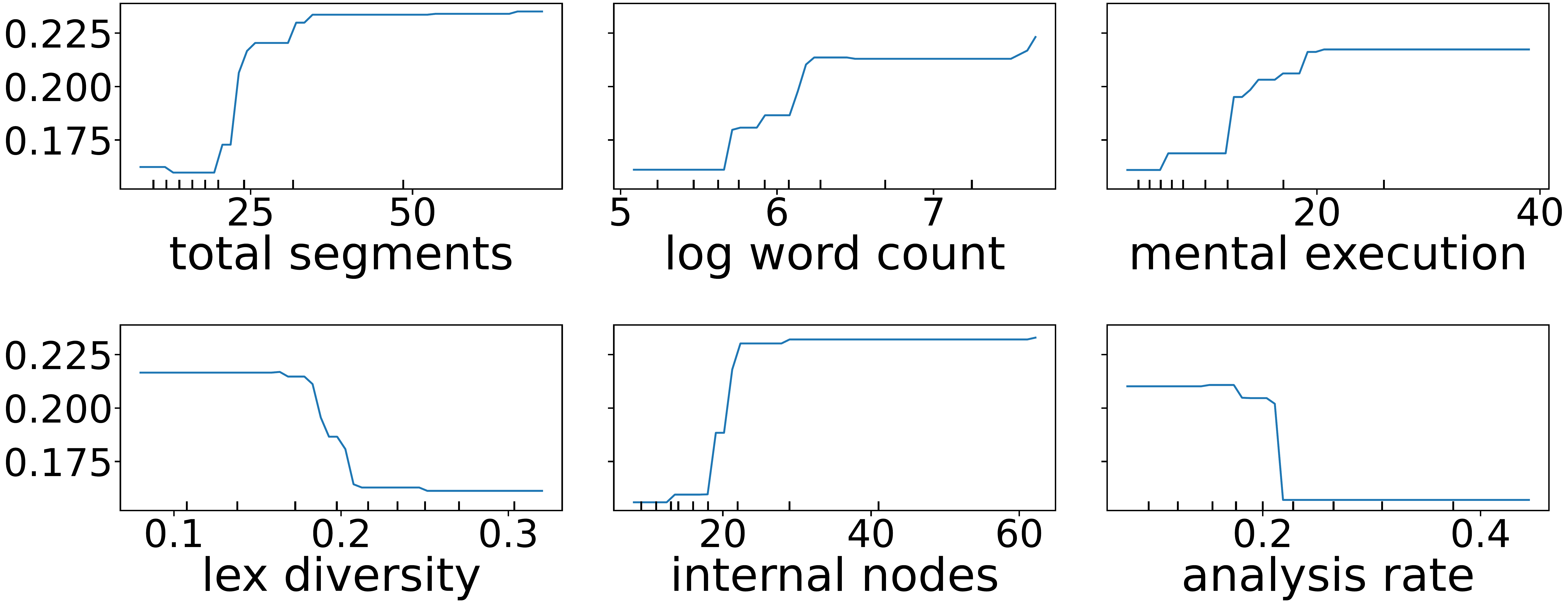}
        \caption{CRUXEval R1}
    \end{subfigure}
    \hfill
    \begin{subfigure}[b]{0.32\textwidth}
        \centering
        \includegraphics[width=\linewidth]{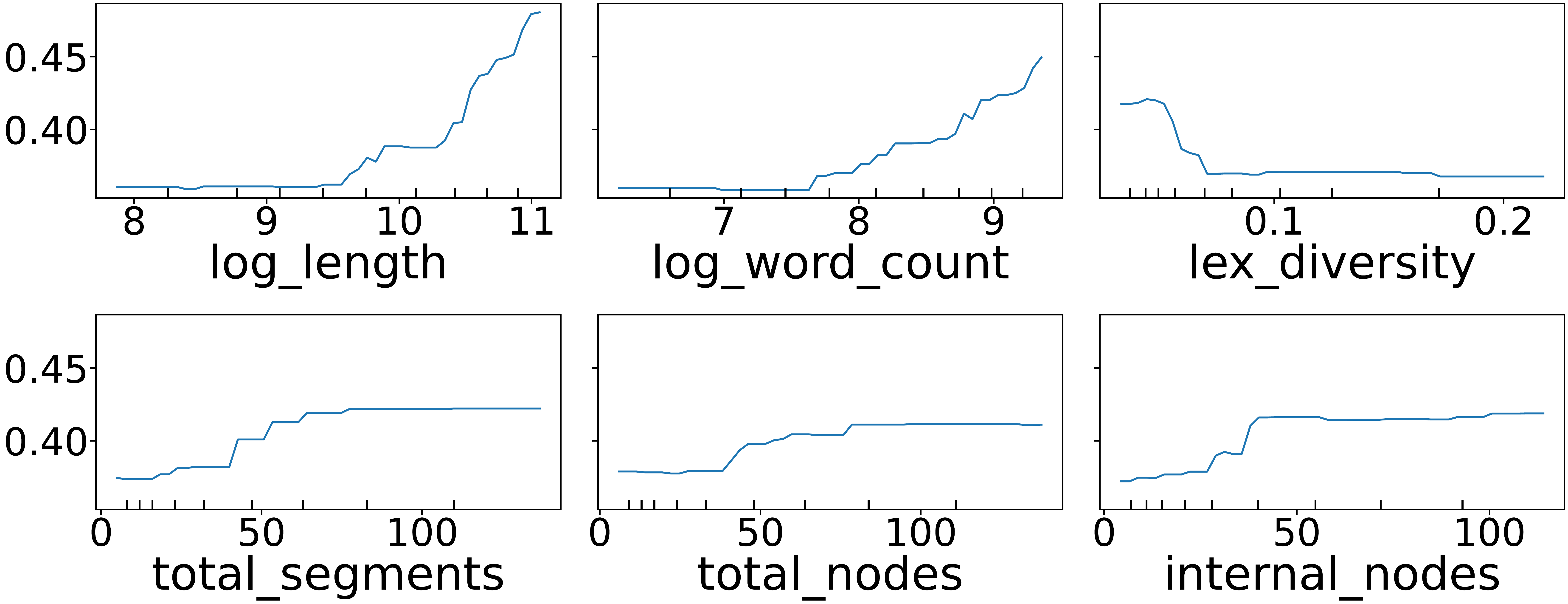}
        \caption{SAFIM R1}
    \end{subfigure}
    \hfill
    \begin{subfigure}[b]{0.32\textwidth}
        \centering
        \includegraphics[width=\linewidth]{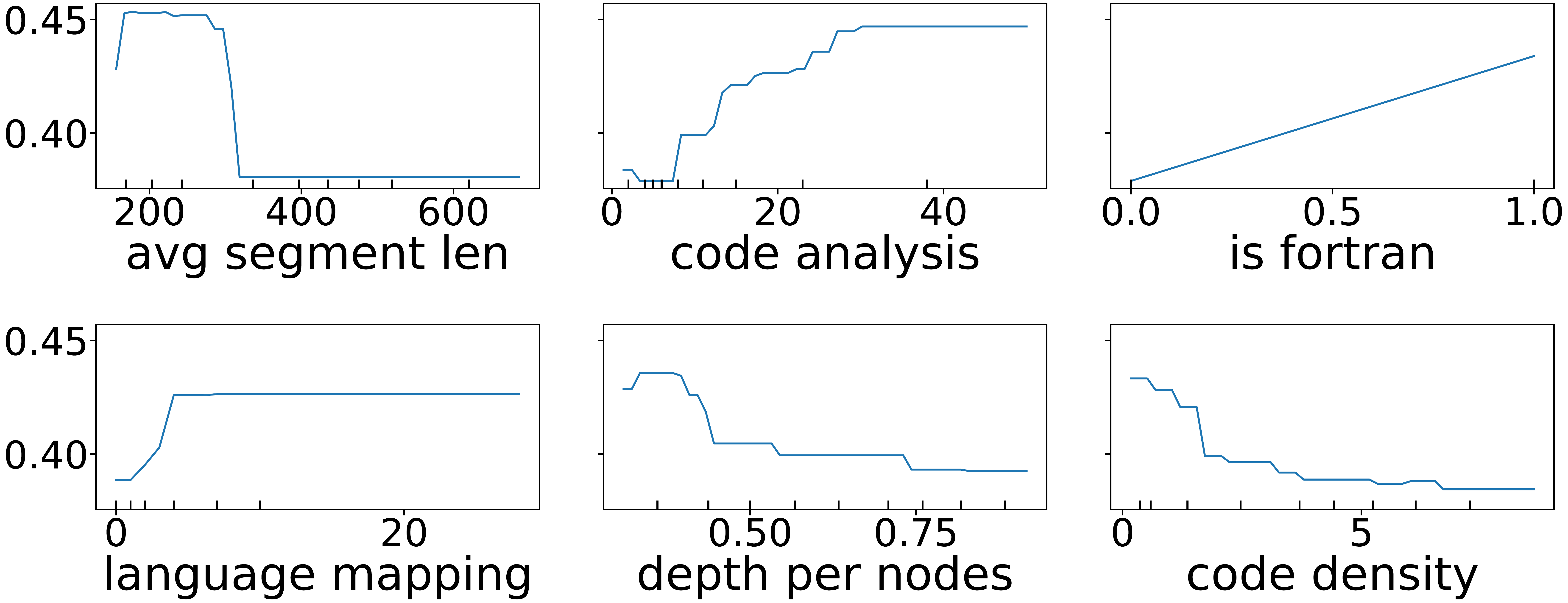}
        \caption{CodeLingua R1}
    \end{subfigure}

    \vspace{1em} 

    \begin{subfigure}[b]{0.32\textwidth}
        \centering
        \includegraphics[width=\linewidth]{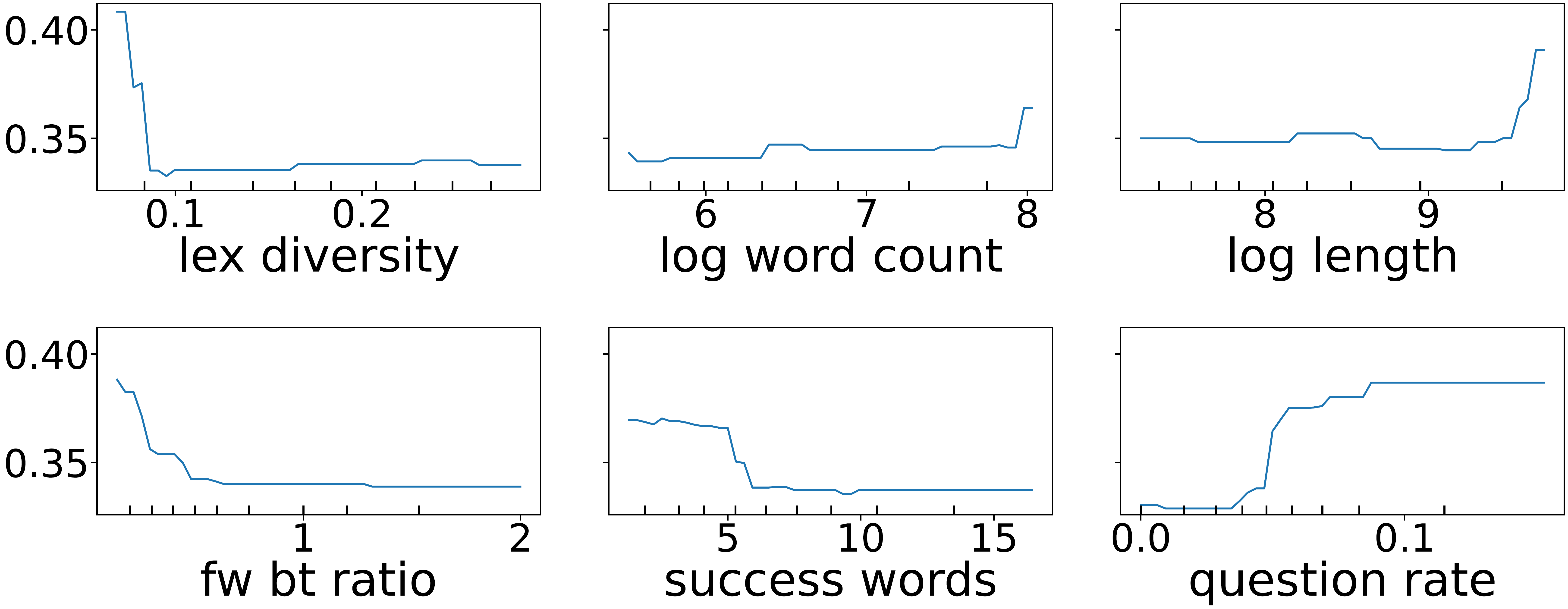}
        \caption{CRUXEval QwQ}
    \end{subfigure}
    \hfill
    \begin{subfigure}[b]{0.32\textwidth}
        \centering
        \includegraphics[width=\linewidth]{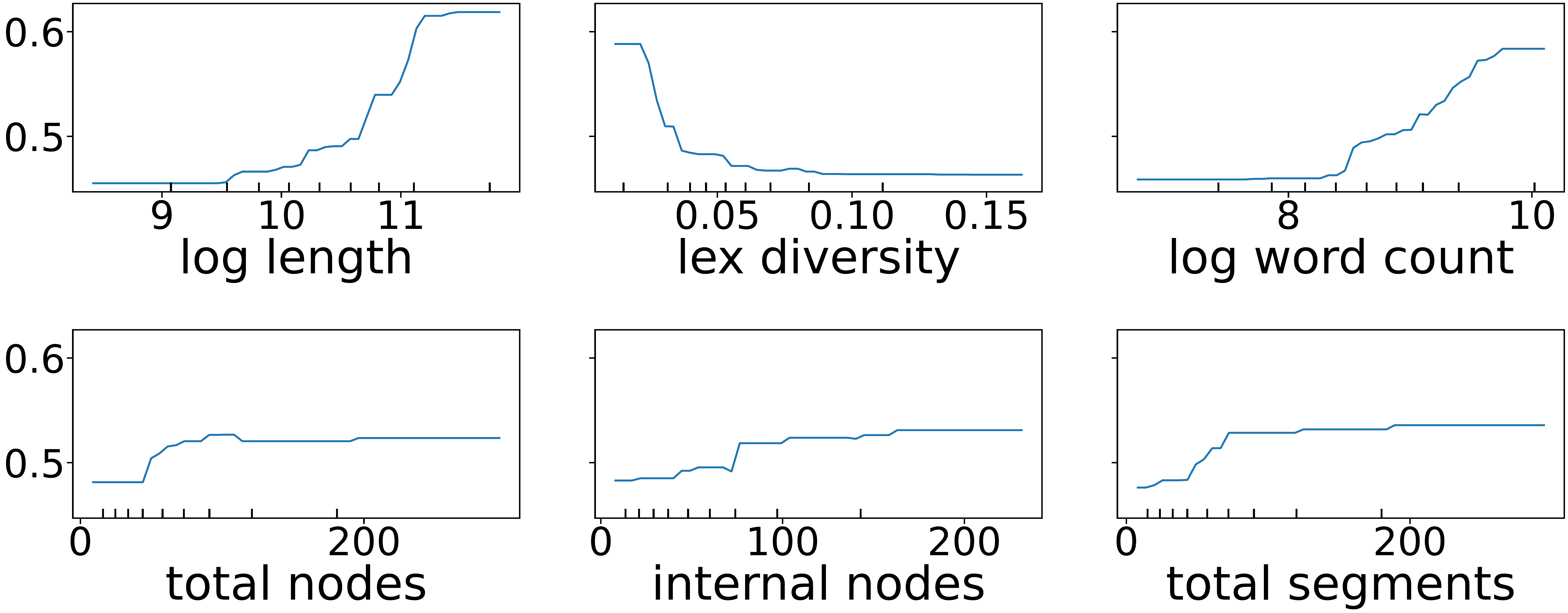}
        \caption{SAFIM QwQ}
    \end{subfigure}
    \hfill
    \begin{subfigure}[b]{0.32\textwidth}
        \centering
        \includegraphics[width=\linewidth]{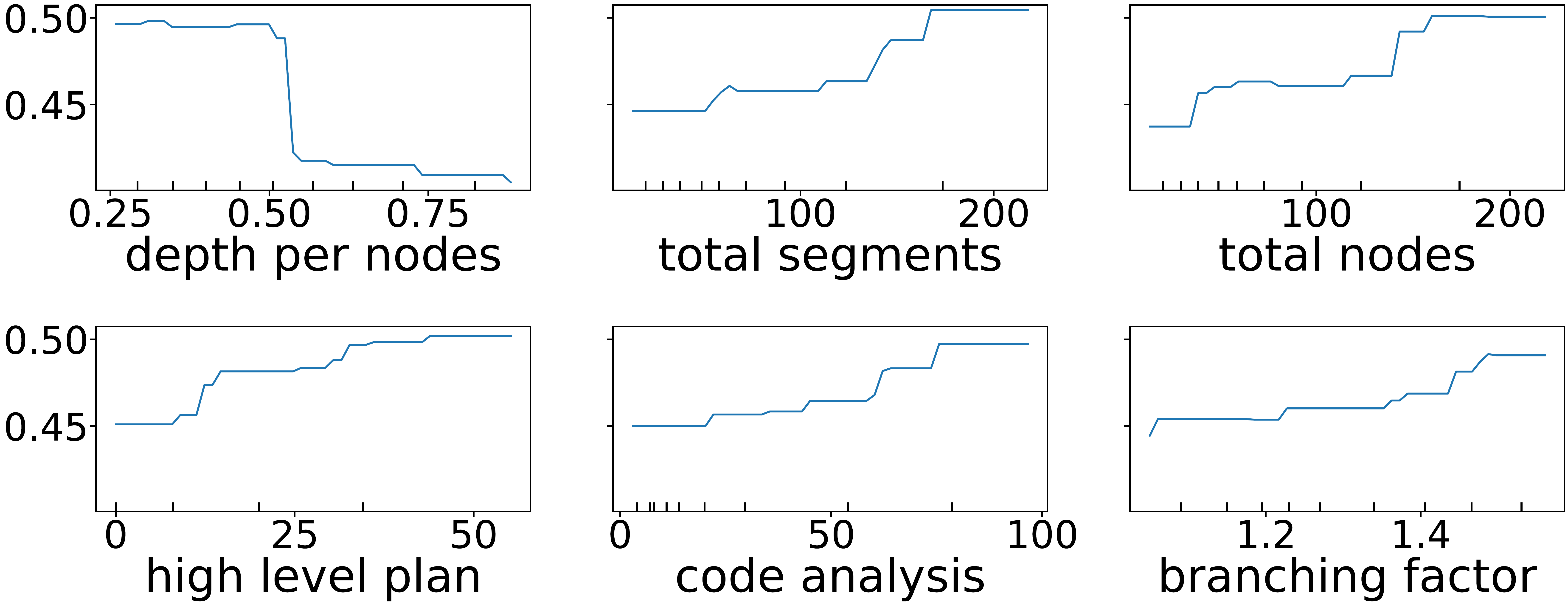}
        \caption{CodeLingua QwQ}
    \end{subfigure}

    \caption{Comparative Partial Dependence Plots across benchmarks (columns) and models (rows). 
}
    \label{fig:pdp_comparison}
\end{figure*}

\section{An ``Oracle'' Classifier}
\label{app:oracle}
This ablation study, described in \cref{sec:oracle_classifier}, trains a single random forest classifier using 15\% of the reasoning traces sampled from each level of CRUXEval, SAFIM, and CodeLingua. The trained model is then evaluated separately on the remaining held-out data from each benchmark.
\begin{figure*}[h]
\centering
\small
\resizebox{\textwidth}{!}{%
\begin{tabular}{llcccccccc}
\toprule
 & & \multicolumn{4}{c}{\textbf{DeepSeek-R1}} & \multicolumn{4}{c}{\textbf{QwQ-32B}} \\
\cmidrule(lr){3-6} \cmidrule(lr){7-10}
\textbf{Dataset} & \textbf{Level} & \textbf{Acc.} & \textbf{W-F1} & \textbf{Inc-F1} & \textbf{Cor-F1} & \textbf{Acc.} & \textbf{W-F1} & \textbf{Inc-F1} & \textbf{Cor-F1} \\ \midrule
CRUXEval & L1 & 0.94 & 0.91 & 0.00 & 0.97 & 0.82 & 0.76 & 0.08 & 0.90 \\
 & L2 & 0.79 & 0.71 & 0.07 & 0.88 & 0.81 & 0.79 & 0.51 & 0.88 \\
 & L3 & 0.87 & 0.85 & 0.56 & 0.92 & 0.78 & 0.76 & 0.46 & 0.86 \\ \midrule
SAFIM & L1 & 0.74 & 0.74 & 0.57 & 0.81 & 0.80 & 0.80 & 0.82 & 0.79 \\
 & L2 & 0.82 & 0.78 & 0.90 & 0.22 & 0.85 & 0.85 & 0.91 & 0.48 \\
 & L3 & 0.81 & 0.76 & 0.89 & 0.07 & 0.88 & 0.89 & 0.93 & 0.47 \\ \midrule
CodeLingua & L1 & 0.73 & 0.71 & 0.31 & 0.83 & 0.68 & 0.68 & 0.55 & 0.75 \\
 & L2 & 0.73 & 0.75 & 0.52 & 0.82 & 0.76 & 0.76 & 0.75 & 0.77 \\
 & L3 & 0.73 & 0.75 & 0.58 & 0.81 & 0.88 & 0.88 & 0.88 & 0.88 \\
\bottomrule
\end{tabular}
}
\caption{Oracle Random Forest classification results across three levels and three datasets, trained and validated on L1, L2, and L3 data for R1 and QwQ. W-F1: Weighted F1; Inc-F1: Incorrect F1; Cor-F1: Correct F1.}
\label{tab:combined_classifier}
\end{figure*}

\clearpage
\section{Feature Correlation}

\begin{figure}[h]
    \centering
    \begin{subfigure}[b]{0.48\textwidth}
        \centering
        \includegraphics[width=\linewidth]
        {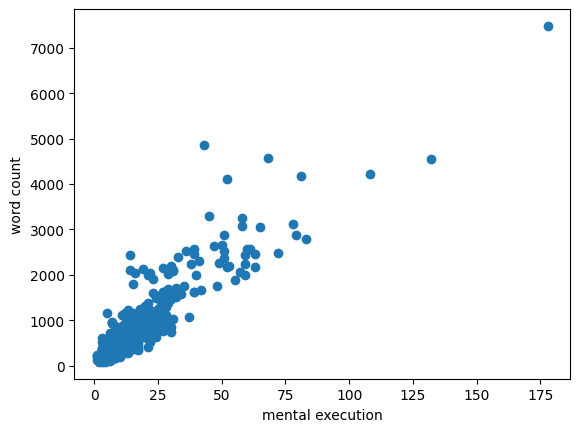}
        \caption{Mental Execution vs. Word Count for CRUXEval Level 1 Reasoning Traces Generated by R1}
        \label{fig:mental_exec_word_count}
    \end{subfigure}
    \hfill 
    \begin{subfigure}[b]{0.48\textwidth}
        \centering
        \includegraphics[width=\linewidth]{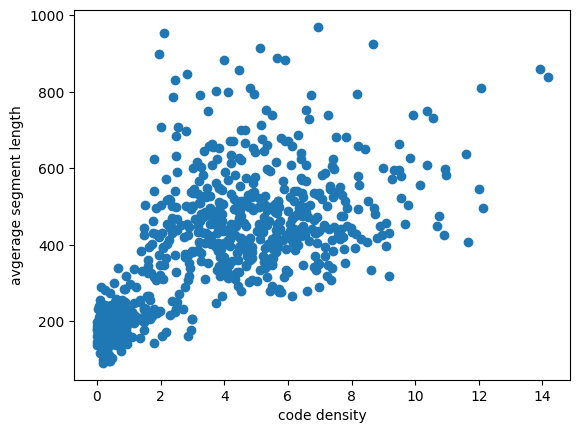}
        \caption{Code Density vs. Average Segment Length for CodeLingua Level 1 Reasoning Traces Generated by R1}
        \label{fig:code_density_avg_seg_len}
    \end{subfigure}
    \caption{Feature Correlation}
\end{figure}

\clearpage
\newpage
\section{Top Features}
\cref{tab:feature_importance} lists the features contributing to the top 50\% of importance for each random forest classifier.
\begin{figure*}[h]
\centering
\small
\begin{tabular}{llc@{\hskip 1.5em}llc}
\toprule
\textbf{Feature (R1)} & \textbf{Imp.} & \textbf{Cum.} & \textbf{Feature (QwQ)} & \textbf{Imp.} & \textbf{Cum.} \\ 
\midrule
\multicolumn{3}{l}{\textbf{CRUXEval}} & \multicolumn{3}{l}{\textbf{CRUXEval}} \\
total\_segments & 8.78\% & 8.78\% & lex\_diversity & 13.72\% & 13.72\% \\
log\_word\_count & 8.55\% & 17.33\% & log\_word\_count & 6.99\% & 20.71\% \\
mental\_execution & 7.79\% & 25.12\% & log\_length & 6.32\% & 27.02\% \\
lex\_diversity & 7.13\% & 32.24\% & fw\_bt\_ratio & 5.94\% & 32.96\% \\
internal\_nodes & 6.69\% & 38.94\% & success\_words & 4.57\% & 37.54\% \\
analysis\_rate & 4.55\% & 43.49\% & question\_rate & 4.22\% & 41.76\% \\
log\_length & 4.05\% & 47.53\% & total\_segments & 4.17\% & 45.93\% \\
total\_nodes & 3.71\% & 51.24\% & mental\_execution & 3.75\% & 49.67\% \\
& & & forward\_rate & 3.74\% & 53.42\% \\ \midrule
\multicolumn{3}{l}{\textbf{SAFIM}} & \multicolumn{3}{l}{\textbf{SAFIM}} \\
log\_length & 10.41\% & 10.41\% & log\_length & 13.05\% & 13.05\% \\
log\_word\_count & 8.66\% & 19.07\% & lex\_diversity & 10.38\% & 23.42\% \\
lex\_diversity & 6.90\% & 25.97\% & log\_word\_count & 9.46\% & 32.89\% \\
total\_segments & 4.59\% & 30.57\% & total\_nodes & 5.47\% & 38.35\% \\
total\_nodes & 4.49\% & 35.06\% & internal\_nodes & 4.46\% & 42.82\% \\
internal\_nodes & 4.17\% & 39.22\% & total\_segments & 4.33\% & 47.15\% \\
internal\_per\_depth & 3.43\% & 42.65\% & max\_segment\_len & 3.75\% & 50.90\% \\
sent\_len\_cv & 3.08\% & 45.74\% & & & \\
depth\_per\_nodes & 2.46\% & 48.20\% & & & \\
code\_analysis & 2.41\% & 50.61\% & & & \\ \midrule
\multicolumn{3}{l}{\textbf{CodeLingua}} & \multicolumn{3}{l}{\textbf{CodeLingua}} \\
avg\_segment\_len & 11.53\% & 11.53\% & depth\_per\_nodes & 10.95\% & 10.95\% \\
code\_analysis & 7.16\% & 18.69\% & total\_segments & 7.26\% & 18.21\% \\
is\_fortran & 6.55\% & 25.24\% & total\_nodes & 7.14\% & 25.35\% \\
language\_mapping & 5.74\% & 30.98\% & high\_level\_plan & 6.72\% & 32.06\% \\
depth\_per\_nodes & 5.41\% & 36.39\% & code\_analysis & 6.69\% & 38.76\% \\
code\_density & 4.96\% & 41.35\% & branching\_factor & 4.81\% & 43.57\% \\
total\_segments & 4.60\% & 45.95\% & internal\_per\_depth & 3.89\% & 47.46\% \\
leaf\_nodes & 3.03\% & 48.98\% & internal\_nodes & 3.62\% & 51.07\% \\
branching\_factor & 2.92\% & 51.91\% & & & \\ \bottomrule
\end{tabular}
\caption{Top features accounting for $>50\%$ of total feature importance across models and datasets.}
\label{tab:feature_importance}
\end{figure*}


\section{Prompts}
\label[appendix]{app:prompts}
\newtcolorbox{snippet}{
  enhanced,
  breakable,
  colback=gray!3,
  colframe=gray!60!black,
  boxrule=0.5pt,
  left=6pt,
  right=6pt,
  top=1pt,
  bottom=1pt,
  sharp corners,
}
We used the following prompts for evaluation for our prediction tasks for the 3 different benchmarks:

\subsection{CRUXEval Prompts}
1) Level 1 Prompt
\begin{tcolorbox}
[top=1pt, bottom=1pt]
Based on the given Python code, which may contain errors, complete the assert statement with the output when executing the code on the given test case. Do not output any extra information, even if the function is incorrect or incomplete.

\{code\}

assert f(\{input\}) ==\\
Return the output of the function without any other things
\end{tcolorbox}

2) Level 2 Prompt
\begin{tcolorbox}
[top=1pt, bottom=1pt]
    Based on the given Python code, which may contain errors, complete the assert statement with the output when executing the code on the given test case. Do not output any extra information, even if the function is incorrect or incomplete.\\
    \# f1\\
    \{f1\_code\}\\
    \# f2\\
    \{f2\_code\}\\
    assert f2(f1(\{input\})) == \\
    Only return the output of the function without any other information and assert statement. If the output is a string, enclose it in single quotes.
\end{tcolorbox}

3) Level 3 Prompt
\begin{tcolorbox}
[top=1pt, bottom=1pt]
    Based on the given Python code, which may contain errors, complete the assert statement with the output when executing the code on the given test case. Do not output any extra information, even if the function is incorrect or incomplete.\\
    \# f1\\
    \{f1\_code\}\\
    \# f2\\
    \{f2\_code\}\\
    \# f3\\
    \{f3\_code\}\\
    assert f3(f2(f1(\{input\}))) == \\
    Only return the output of the function without any other information and assert statement. If the output is a string, enclose it in single quotes.
\end{tcolorbox}

\subsection{SAFIM Prompts}
1) Level 1 Prompt
\begin{tcolorbox}
[top=1pt, bottom=1pt]
You will be given code with missing lines or blocks that you must fill in.\\
Output only the missing code so that the program will run correctly. 
Output the missing code as plain text, NOT as markdown code. 
Do NOT output the entire program or any additional information.

\{code\}
\end{tcolorbox}

2) Level 2 Prompt
\begin{tcolorbox}
[top=1pt, bottom=1pt]
You will be given program1.py and program2.py with missing lines or blocks that you must fill in.\\
Output only the missing code so that "python3 program1.py \textbar{ } program2.py" runs correctly.\\
Output the missing code, NOT as markdown code. \\
Do NOT output the entire program or explanations or any additional information.\\

\# Output Format\\
Return json with the following structure:\\
\{\{“program1”: “completion”, “program2”: “completion”\}\}

\# program1.py\\
\{f1 problem description $+$ code\}

\# program2.py\\
\{f2 problem description $+$ code\}
\end{tcolorbox}

3) Level 3 Prompt
\begin{tcolorbox}
[top=1pt, bottom=1pt]
You will be given program1.py, program2.py and program3.py with missing lines or blocks that you must fill in.\\
Output only the missing code so that "python3 program1.py \textbar{ } program2.py \textbar{ } program3.py" runs correctly.\\
Output the missing code, NOT as markdown code. \\
Do NOT output the entire program or explanations or any additional information.\\

\# Output Format\\
Return json with the following structure:\\
\{\{“program1”: “completion”, “program2”: “completion”, “program3”: “completion”\}\}

\# program1.py\\
\{f1 problem description $+$ code\}

\# program2.py\\
\{f2 problem description $+$ code\}

\# program3.py\\
\{f3 problem description $+$ code\}
\end{tcolorbox}

\subsection{Codelingua Prompts}
1) Level 1 DeepSeek-R1 Prompt
\begin{tcolorbox}
[top=1pt, bottom=1pt]
You are an AI programming assistant, utilizing the DeepSeek Coder model, developed by DeepSeek Company, and you only answer questions related to computer science. For politically sensitive questions, security and privacy issues, and other non-computer science questions, you will refuse to answer.\\
\#\#\# Instruction:\\
Translate the following \{source\_language\} code to \{target\_language\}.\\

\{source\_language\}\\
\{code\}\\

\#\#\# Response:
\end{tcolorbox}

2) Level 1 QwQ Prompt
\begin{tcolorbox}
[top=1pt, bottom=1pt]
Translate the following code from \{source\_language\} to \{target\_language\}: \\
\{code\}
\end{tcolorbox}

3) Higher Level Prompt

\begin{tcolorbox}
[top=1pt, bottom=1pt]
You are given a set of \{source\_language\} programs that are meant to be executed in sequence, where the output of each program is used as the input to the next.

Translate the *entire sequence* into a single \{target\_language\} program that reproduces the same behavior.\\
    - Only the first block should handle reading input.\\
    - Only the last block should handle producing output.\\
    - Intermediate steps should process data *without I/O*.\\
    - Do not insert any additional print/read statements in the middle.\\
    Ensure that the final \{target\_lang\} program behaves identically to the original chain when run on the same input.\\
    \{source\_code\}
\end{tcolorbox}

\subsection{Semantic Labeling Prompt}
\begin{tcolorbox}
    You are an expert in reasoning trace analysis.\\

Classify the following reasoning segment into one of these strategies:\\

- "code\_analysis": actively analyzing what the original code does\\
- "mental\_execution": simulating input/output behavior\\
- "test\_generation": proposing test cases or edge conditions\\
- "bug\_fixing": describing or fixing syntax/logic errors\\
- "language\_mapping": translating code constructs across languages\\
- "high\_level\_plan": outlining the approach or plan without executing it\\
- "empty": no meaningful reasoning\\

Only return a JSON like:\\
```json\\
\{\{"type": "..."\}\}\\
Segment:\\
\{segment\}
\end{tcolorbox}

\subsection{Tree Generation Prompt}
\begin{tcolorbox}
[top=1pt, bottom=1pt]
You are building a reasoning tree from segments of thought. Each segment expresses a part of a reasoning process.\\

Here are previous thoughts in the structure:\\
\{tree\_so\_far\}\\

Now consider this new segment:\\
```\\
\{new\_segment\}\\
```\\

Decide where to attach this new segment in the tree. You must use one of the three relation types defined below.\\

Relation Categories\\
- Continuation: The new segment directly builds upon the parent's idea. It can do this by:\\
    1. Adding more details, explanations, or examples.\\
    2. Providing evidence or justification.\\
    3. Refining or clarifying the parent's point.\\
- Contrast: The new segment proposes a different, alternative, or opposing idea compared to the parent node.\\
- Rephrase: The new segment expresses the exact same core idea as the parent but in different words.\\

Return your answer in this JSON format:\\
```json\\
\{\{\\
  "parent\_id": "thought\_X",  // id of parent thought, or "root" for top-level\\
  "relation": "Continuation" \textbar{} "Contrast" \textbar{} "Rephrase"\\
\}\}
\end{tcolorbox}

\subsection{Intervention Prompt}
\label{sec:intervention_prompt}
Based on the key predictive features identified in \cref{fig:pdp_comparison}, we construct a structured prompt that summarizes common failure signals and prepend it to each problem before retry.
\begin{tcolorbox}
    Additional guidance: prior analysis of successful runs on this benchmark suggests your reasoning style on similar problems tends to be more compact than on failed attempts. Keep internal reasoning concise: simulate f on the given input in few steps, avoid redundant backtracking, then output only the literal result as instructed.\\
    You will be given code with missing lines or blocks that you must fill in.\\
    Output only the missing code so that the program will run correctly. Output the missing code as plain text, NOT as markdown code. Do NOT output the entire program or any additional information.
\end{tcolorbox}

\section{Qualitative Examples}

\subsection{CRUXEval Example}

\begin{longtable}{r l p{12cm}}
\hline
\textbf{\#} & \textbf{Label} & \textbf{Reasoning Segment} \\ \hline
\endfirsthead

\hline
\textbf{\#} & \textbf{Label} & \textbf{Reasoning Segment (Continued)} \\ \hline
\endhead

\hline
\multicolumn{3}{r}{\textit{Continued on next page}} \\
\endfoot

\hline
\endfoot
1 & empty & We are given two functions: f1 and f2. \\ \hline
2 & mental\_execution & We are to compute: assert f2(f1('\"Makers of a Statement\"')) == ? \\ 
  & & First, let's compute f1('\"Makers of a Statement\"'): s = '\"Makers of a Statement\"' $\rightarrow$ s = s.replace('\"', '') removes all double quotes, so becomes: 'Makers of a Statement'. lst = list(s) $\rightarrow$ ['M','a','k','e','r','s',' ','o','f',' ','a',' ','S','t','a','t','e','m','e','n','t']. col = 0, count = 1. while col < len(lst) and lst[col] in \texttt{.:,}: at col=0, lst[0] is 'M' $\rightarrow$ not in \texttt{.:,} $\rightarrow$ so the loop condition fails. \\ \hline
3 & mental\_execution & Therefore, we break out of the loop without entering. Then return s[col+count:] $\rightarrow$ s[0+1:] $\rightarrow$ s[1:] $\rightarrow$ 'akers of a Statement'. \\ 
  & & Now, we call f2 on the result: f2('akers of a Statement'). In f2: i = 0. while i < len(text) and text[i].isspace(): text[0] is 'a' $\rightarrow$ not a space $\rightarrow$ condition fails. Break loop. i == len(text)? i=0 and len(text)=20 $\rightarrow$ false. Return 'no'. \\ 
  & & Note: f1 has an error: \texttt{count = ls[col] + 1}. Should be 'lst'. But in this test case, the loop did not run, so the line was never executed. \\ \hline
4 & code\_analysis & Therefore, the error did not manifest. Examining f1: The loop condition \texttt{while col < len(lst) and lst[col] in ".:,"} skips execution because the first character is 'M'. The result of f1 is 'akers of a Statement'. f2 returns 'no'. \\ \hline
5 & bug\_fixing & Note: the problem states Python code may contain errors. In this specific test case, the index error in f1 (\texttt{ls} vs \texttt{lst}) is not encountered. \\ \hline
6 & empty & Therefore, the final answer is 'no'. \\ \hline

\caption{Complete Reasoning Trace of a CRUXEval L2 Example with Segmented Labels} 
\end{longtable}

\subsection{SAFIM Example}

\begin{longtable}{r l p{12cm}}
\hline
\textbf{\#} & \textbf{Label} & \textbf{Reasoning Segment} \\ \hline
\endfirsthead

\hline
\textbf{\#} & \textbf{Label} & \textbf{Reasoning Segment (Continued)} \\ \hline
\endhead

\hline
\multicolumn{3}{r}{\textit{Continued on next page}} \\
\endfoot

\hline
\endfoot

1 & high\_level\_plan & We are given three programs with missing parts. We must output the missing code for each program in a JSON format. \\ \hline
2 & code\_analysis & For program1.py: The problem is about counting the number of valid permutations $c$ that can be formed by choosing at each index $i$ either $a\_i$ or $b\_i$, such that $c$ is a permutation and matches the constraints given by $d$. The provided code builds a graph and then tries to traverse cycles. \\ \hline
3 & code\_analysis & Approach: We note that the problem can be modeled as a graph where each node (number from 1 to $n$) has an edge to the two numbers that are at the same index in $a$ and $b$? Actually, the graph is built as: \texttt{g = \{ a[i]: [b[i], c[i]] for i in range(n) \}}. \\ \hline
4 & code\_analysis & Because if we choose $a[i]$ at this index, then the value $a[i]$ must appear and we cannot choose the same value again. But note: the constraint $d$ (called $c$ in code) might fix a value. \\ \hline
5 & high\_level\_plan & The idea is to form a graph where each node (value) has one outgoing edge to the other value that appears at the same index. At index $i$, we have values $a[i]$ and $b[i]$. The key observation is that the constraints and the choices form cycles. \\ \hline
6 & high\_level\_plan & In each cycle, we have two choices for the entire cycle (like a circular linked list) unless the cycle is broken by a constraint. \\ \hline
7 & code\_analysis & Build a graph where each index $i$ has an edge from $a[i]$ to $b[i]$ and stores constraint $d[i]$. Traverse unvisited nodes to find connected components (cycles). The answer is $2^{\text{free cycles}} \pmod{10^9+7}$. \\ \hline
8 & mental\_execution & However, note: the constraint might be either $a[i]$ or $b[i]$. If the constraint is set, then we know which one to choose at that index, and that breaks the cycle into a chain? \\ \hline
9 & code\_analysis & Actually, the constraint at one index forces the choice at that index, and then the rest of the cycle is forced? So the entire cycle has only one possibility. Disjoint union of cycles. \\ \hline
10 & code\_analysis & However, note: the graph is built for every $a[i]$ and we have $n$ keys. So every node is a key. But what about the constraint? We must check every node in the cycle? \\ \hline
11 & bug\_fixing & Actually, in the current code, we are only checking the constraint at the starting node? We must check every node in the cycle. Traverse the entire cycle and set a flag \texttt{free} to True initially. \\ \hline
12 & code\_analysis & If we find at least one constraint, then the entire cycle is forced (so \texttt{free} becomes False). The constraint $d[i]$ is associated with the index $i$, and we are storing it by the value $a[i]$. \\ \hline
13 & code\_analysis & However, when we traverse the cycle, we are going from $a[i]$ to $b[i]$ and then we use $b[i]$ as the next key? But the next key is the $a[j]$ for the next index $j$? \\ \hline
14 & code\_analysis & Actually, the graph is built so that for each $a[i]$ we have an edge to $b[i]$. When at node $u$, we are at value $u$ which is the $a[i]$ for some index $i$. If \texttt{free} and \texttt{cycle\_size} > 1, we have 2 choices. \\ \hline
15 & code\_analysis & Recall that a permutation is an array consisting of $n$ distinct integers. So the values in $a$ are distinct, and the values in $b$ are distinct. But at the same index, $a\_i$ and $b\_i$ could be the same? \\ \hline
16 & mental\_execution & If $a_i = b_i$, the edge is a self-loop. We check if \texttt{need} and \texttt{cycle\_size} > 1. For a self-loop, we have no freedom. If any constraint is non-zero, the cycle is not free. \\ \hline
17 & code\_analysis & We break the while loop when we hit a visited node. The condition \texttt{while not visi[g[u][0]]} checks the next node. This does not check the constraint at the last node? \\ \hline
18 & bug\_fixing & Because when we are at the last node, we set $u$ to the next (first node) and break because it's visited. We need to traverse the entire cycle to process every node's constraint. \\ \hline
19 & code\_analysis & This condition is on the next node? But we haven't visited the next node? If \texttt{g[current][1] != 0}, then \texttt{need = False}. \\ \hline
20 & code\_analysis & The code starts at $u$, but we do not mark $u$ as visited? If we move to the next node and mark it visited, what happens when we start at $u$ again? Iterate $u$ from 1 to $n$. \\ \hline
21 & mental\_execution & Revised plan: start at $u$, mark it visited, check constraint at $u$. Move to \texttt{g[u][0]} and mark it visited. Process every node in the cycle sequence $u_0 \rightarrow u_1 \dots \rightarrow u_{k-1} \rightarrow u_0$. \\ \hline
22 & mental\_execution & We processed it at the first step. But then when we were at $u_{k-1}$, we set $u$ to $u_0$, and then we break? The constraint at $u_0$ was already processed at the first step. \\ \hline
23 & mental\_execution & The constraint at $u_0$ is stored at the node $u_0$, and we processed it when $u$ was $u_0$. So we have processed every node? \\ \hline
24 & mental\_execution & Starting node $u_0$ is not visited initially? Then enter the loop only if next node is not visited. If cycle length is 1 ($u_0 \rightarrow u_0$), \texttt{cycle\_size} remains 1 and we skip. \\ \hline
25 & bug\_fixing & We must set \texttt{need} to False if we encounter a constraint at any node. The missing code for program1.py is: \texttt{if g[u][1] != 0: need = False}. \\ \hline
26 & bug\_fixing & However, the code has a bug: we start at $u$, but do not mark $u$ as visited? If $u$ is already visited, we should skip the for loop body. \\ \hline
27 & code\_analysis & For loop condition \texttt{for u in range(1, n+1)}: if the next node of $u$ is not visited, enter. If $u$ is visited, skip the while loop? \\ \hline
28 & code\_analysis & If the next node of $u$ is not visited, then we enter? But what if $u$ is visited? Then we skip the while loop? But we haven't processed $u$? \\ \hline
29 & mental\_execution & Alternatively, we can skip visited nodes? We iterate $u$ from 1 to $n$. Mark every node in the cycle as visited. Ensure starting node $u$ is marked correctly. \\ \hline
30 & bug\_fixing & Revised: \texttt{if visi[u]: continue}. Need to process the constraint at the starting node $u$. Set \texttt{cycle\_size = 1} because we processed one node. \\ \hline
31 & mental\_execution & While True loop: if \texttt{visi[current]}, break. Otherwise, mark visited and check \texttt{g[current][1]}. This traverses the entire component cycle. \\ \hline
32 & code\_analysis & In program1.py, the TODO is inside the while loop. We add: \texttt{if g[u][1] != 0: need = False}. Indentation must match the outer block. \\ \hline
33 & code\_analysis & Actually, the choice at index $i$: if we choose $a\_i$, then we use the value $a\_i$ and then the next node in the cycle is $b\_i$? \\ \hline
34 & code\_analysis & if we choose $b\_i$, then we use the value $b\_i$ and then the next node in the cycle is $a\_i$? The graph is built as: from $a\_i$ we go to $b\_i$. \\ \hline
35 & code\_analysis & If any constraint is set, the entire cycle is forced. If \texttt{g[u][1] != 0}, then \texttt{need = False}. If constraint is 0, leave \texttt{need} unchanged. \\ \hline
36 & code\_analysis & For program2.py: Remove all occurrences of $t$ in $s$ with minimum moves and count optimal sequences. Dynamic programming state \texttt{d[j][i]} for substring \texttt{s[i:j+1]}. \\ \hline
37 & code\_analysis & The recurrence: consider removing an occurrence at $k$. Split into left part \texttt{s[i:k-1]} and right part \texttt{s[k+lt:j]}. Base case empty substring is \texttt{[0,1]}. \\ \hline
38 & bug\_fixing & Total moves = $l[0] + r[0] + 1$. Total ways = $l[1] \times r[1]$. Current code adds only $r[1]$ and has a suspicious \texttt{if l[0] == 0} condition. \\ \hline
39 & bug\_fixing & The recurrence should apply regardless of \texttt{l[0]}. But we are only allowed to fill the TODO inside the \texttt{if l[0] == 0} block. \\ \hline
40 & code\_analysis & If \texttt{l[0] == 0}, then \texttt{l[1]} is likely 1. So \texttt{new\_ways = r[1]}. The TODO should set \texttt{tt = [l[0] + r[0] + 1, r[1]]}. \\ \hline
41 & high\_level\_plan & Constraints are small ($|s| \le 500$). We can use prefix DP \texttt{dp[i] = [min\_moves, count]} for prefix \texttt{s[0:i]}. \\ \hline
42 & high\_level\_plan & Alternatively, interpret as contiguous removals? Problem says replace with dots. Dots are placeholders and don't affect future matches. \\ \hline
43 & code\_analysis & If we skip character, \texttt{dp[i+1]} updates from \texttt{dp[i]}. If we remove occurrence ending at $i$, update with $+1$ move. \\ \hline
44 & code\_analysis & Actually, can only remove contiguous? After removal, string is \texttt{s[0:k] + '.'*lt + s[k+lt:]}. \\ \hline
45 & code\_analysis & replace with dots. The dots are not letters, so they cannot form part of a future occurrence. \\ \hline
46 & mental\_execution & Try every occurrence ending at $i$. This starts at $i-lt+1$. Non-overlapping occurrences. This is complex but standard interval DP. \\ \hline
47 & code\_analysis & Iterate $j$ and check if \texttt{s[j:j+lt] == t}. If no overlap, update. Missing code for program2.py: \texttt{tt = [l[0] + r[0] + 1, r[1]]}. \\ \hline
48 & bug\_fixing & If left substring is non-empty, \texttt{d[k-1][i]} having 0 moves means it was already empty? If \texttt{l[0]==0}, then \texttt{l[1]=1}. So TODO is \texttt{tt = [l[0] + r[0] + 1, r[1]]}. \\ \hline
49 & code\_analysis & For program3.py: Given array $C$, find binary array $A$ such that $\sum B_k = C$, where $B_k$ is $A$ with first $k$ elements sorted. \\ \hline
50 & mental\_execution & For $k > i$: position $i$ is 1 iff number of ones in first $k \ge k - i$. \\ \hline
51 & mental\_execution & Contribution: $C_i = i \cdot a_i + \sum_{k=i+1}^n [1 \text{ if } P_k \ge k-i \text{ else } 0]$. Recover array from left to right. \\ \hline
52 & mental\_execution & Total ones $T = \text{sum}(C) / n$. If $C_{n-1} == n$, then $a_{n-1} = 1$, else 0. This works for small $n$. \\ \hline
53 & high\_level\_plan & Given the provided code has an extra 1 at the end and a loop, we do a greedy approach. Initialization: \texttt{a = [1 if x else 0 for x in c] + [1]}. \\ \hline
54 & high\_level\_plan & Use a Fenwick tree or difference array to track active ones. Constraints are 200,000. $C$ is a sum of suffix range updates for each $k$. \\ \hline
55 & high\_level\_plan & For a fixed $k$, it contributes 1 to every $i \ge$ number of zeros in the first $k$. $C\_i = i \cdot a\_i + \text{suffix sum of diffs}$. \\ \hline
56 & mental\_execution & Example: $c = [2,4,2,4], n=4$. $a=[1,1,0,1]$ works. $a_0 = 1$ because $2>0$, $a_1=1$ because $4>1$, $a_2=0$ because $2>2$ is false. \\ \hline
57 & mental\_execution & "It is guaranteed that a valid array $A$ exists". So $T$ must be integer. If $C=[6,5,2,1,4,3]$, $T=21/6=3.5$ (typo in example note). \\ \hline
58 & mental\_execution & In the third test case, permutations are $c$ arrays. Binary array $a$ is not provided. \\ \hline
59 & test\_generation & Input $C$ contains $n$ integers. $T = 21/6$. Mistake in the example note regarding $C$ values. \\ \hline
60 & mental\_execution & In the third test case, permutations $c$ can be made using the process. These are the result of sorting, not the input binary array. \\ \hline
61 & code\_analysis & Input is one array $C$. Output is binary array $A$. Logic $a[i] = 1$ if $c[i] > i$ else 0 works for first and fifth test cases. \\ \hline
62 & code\_analysis & "In the third test case, there are 2 distinct permutation". These are permutations $c$. $A$ is the binary source. \\ \hline
63 & bug\_fixing & Missing code for program3.py: \texttt{if x > i: a[i] = 1 else: a[i] = 0}. The extra 1 at the end of $a$ is removed by \texttt{a[:-1]}. \\ \hline
64 & mental\_execution & Loop sets $a[i]$ for $i$ in range $n$. Output $a[:-1]$ matches length $n$. Condition $x > i$ is the simplest valid heuristic. \\ \hline
65 & mental\_execution & Final JSON structure: program1 (\texttt{need=False}), program2 (\texttt{tt} assignment), program3 (\texttt{a[i]} assignment). \\ \hline
\caption{Complete Reasoning Trace of a SAFIM L3 Example with Segmented Labels} \\
\end{longtable}

\subsection{CodeLingua Example}
\label[appendix]{app:full_trace}

\begin{longtable}{r l p{12cm}}
\hline
\textbf{\#} & \textbf{Label} & \textbf{Reasoning Segment} \\ \hline
\endfirsthead

\hline
\textbf{\#} & \textbf{Label} & \textbf{Reasoning Segment (Continued)} \\ \hline
\endhead

\hline
\multicolumn{3}{r}{\textit{Continued on next page}} \\
\endfoot

\hline
\endfoot

1 & \texttt{lang\_mapping} & Okay, so I need to translate this Java code into Fortran. \\ \hline
2 & \texttt{code\_analysis} & Let me start by understanding what the Java code does. \\ \hline
3 & \texttt{code\_analysis} & The main part seems to be handling a range of integers from 0 to n-1, stored in a TreeSet called 'left'. \\ \hline
4 & \texttt{high\_level\_plan} & For each query, they take a range [l, r] and a winner 'win'. \\ \hline
5 & \texttt{code\_analysis} & Then, they iterate over all elements in 'left' that are between l and r, remove them, set their answer to 'win', but then set the winner's own position back to 0 and add it back to 'left'. \\ \hline
6 & \texttt{code\_analysis} & Wait, the Java code has a loop where for each query, they process the current l to r range. They find the ceiling of l in the TreeSet, which is the smallest element $\geq$ l. Then, while that element is $\leq$ r, they remove it from the set, set its answer to 'win'. But then, after processing all those elements, they set answer[win-1] to 0 and add win-1 back to the set. \\ \hline
7 & \texttt{code\_analysis} & Hmm, so the winner's position is kept in the set, but others in the range are removed and their answers are set to 'win'. So the TreeSet is being used to keep track of the remaining positions that haven't been processed yet. \\ \hline
8 & \texttt{high\_level\_plan} & Each query affects a range, removing those positions (except the winner) and marking their answer as the winner. \\ \hline
9 & \texttt{high\_level\_plan} & The TreeSet is being used to efficiently find and iterate over elements in a range. \\ \hline
10 & \texttt{lang\_mapping} & In Java, ceiling() and higher() methods are used. \\ \hline
11 & \texttt{high\_level\_plan} & One possible approach is to represent the remaining elements in a sorted list or array. But maintaining a sorted structure efficiently in Fortran could be challenging. \\ \hline
12 & \texttt{high\_level\_plan} & Alternatively, maybe we can use a binary search approach on an array that's kept sorted. \\ \hline
13 & \texttt{high\_level\_plan} & Alternatively, since the initial set is 0 to n-1, and elements are being removed and added back (only the winner's position is added back each time), perhaps we can track the elements using a boolean array where each index indicates whether it's present in the 'left' set. But then, how do we efficiently find the next element $\geq$ l and $\leq$ r? \\ \hline
14 & \texttt{code\_analysis} & Hmm, maybe a bitmask or an array of flags. But checking each element from l to r each time could be O(n) per query, which might be acceptable if the constraints are small. But the original Java code uses a TreeSet, which is O(log n) per operation. \\ \hline
15 & \texttt{high\_level\_plan} & The problem statement mentions codeforces\_356\_A, which I think is a programming competition problem. So perhaps the constraints are tight, and we need an efficient solution. \\ \hline
16 & \texttt{high\_level\_plan} & Wait, but in Fortran, we don't have data structures like TreeSet. So perhaps the best approach is to simulate the TreeSet using an array that's kept sorted, and use binary search to find the required elements. \\ \hline
17 & \texttt{high\_level\_plan} & Let's think about how to manage this. \\ \hline
18 & \texttt{high\_level\_plan} & The initial left set is all integers from 0 to n-1. So we can start with a sorted array. \\ \hline
19 & \texttt{high\_level\_plan} & For each query, we need to find all elements in [l, r], set their answer to win, remove them from the array, then add back the win-1 position (if it was removed) and set its answer to 0. But how to efficiently find and remove elements in a range [l, r] from a sorted array. \\ \hline
20 & \texttt{high\_level\_plan} & Let's see: Suppose the array is sorted. We can find the first index where element $\geq$ l using binary search. Then, iterate through the elements from that index until we find elements $\leq$ r. \\ \hline
21 & \texttt{high\_level\_plan} & All those elements need to be removed. For each such element, set answer[element] = win. Then remove them from the array. \\ \hline
22 & \texttt{high\_level\_plan} & After that, check if win-1 is in the array. \\ \hline
23 & \texttt{code\_analysis} & Wait, no: in the Java code, after removing the elements, they set answer[win-1] to 0 and add it back to the set. So even if win-1 was not in the range [l, r], but perhaps it's possible. \\ \hline
24 & \texttt{code\_analysis} & Wait, in the Java code, during each query, after processing the elements in [l, r], they set answer[win-1] to 0 and add it to the set. So regardless of whether win-1 was in the range, they are adding it back. But wait, maybe the processing in the query's loop could have removed win-1. \\ \hline
25 & \texttt{mental\_exec} & For example, if the [l, r] includes win-1, then during processing, it would have been removed from the set. \\ \hline
26 & \texttt{code\_analysis} & Then, after that, setting answer[win-1] to 0 and adding it back. So the Java code ensures that the winner's position is always in the set after each query. \\ \midrule

\multicolumn{3}{c}{\textit{[... 162 Segments Omitted for Brevity ...]} } \\ \midrule

189 & \texttt{code\_analysis} & For example, when after processing, the left array is empty. \\ \hline
190 & \texttt{mental\_exec} & Then, inserting win\_0based would create a new array with one element. \\ \hline
191 & \texttt{high\_level\_plan} & Also, when removing elements, if the indices are out of bounds, the code should handle that. \\ \hline
192 & \texttt{code\_analysis} & Now, in the code for processing the query: [Code snippet involving lower\_bound and upper\_bound calls]. This part removes the elements from left. \\ \hline
193 & \texttt{lang\_mapping} & The code uses array slicing and concatenation. But in Fortran, array slicing is allowed, and array concatenation using the [ ... ] syntax is possible. But when left is empty (size 0), the code may have issues. Need to handle that. \\ \hline
194 & \texttt{high\_level\_plan} & Now, the final step is to print the answer array. \\ \hline
195 & \texttt{code\_analysis} & The Java code appends each answer as a string. \\ \hline
196 & \texttt{code\_analysis} & In Fortran, we can loop through the answer array and print each element, space-separated. But the Java code's output is a string with each answer concatenated. Wait, no. [Analysis of StringBuilder and output formatting]. \\ \hline
197 & \texttt{lang\_mapping} & In Fortran, this can be done using a format that writes all elements without spaces. For example: [Code snippet using do i = 1, n and write statement with advance='no']. \\ \hline
198 & \texttt{code\_analysis} & After all elements are written, a new line is printed. \\ \hline
199 & \texttt{empty} & So, the final code would have this part. Putting it all together. \\ \hline
200 & \texttt{mental\_exec} & Now, let's test this code with a small example. [Step-through of n=3, q=1, l=0, r=2, win=2]. \\ \hline
201 & \texttt{mental\_exec} & Then, set answer[1] (win-1 is 1) to 0, and add 1 back to left. So answer array is [2, 0, 2]. So output is "202". \\ \hline
202 & \texttt{mental\_exec} & In Fortran: answer array is 1-based. answer(1) = 2, answer(2) =0, answer(3) =2. So the output would be "202". Another test case. [Step-through of 5 2 input]. \\ \hline
203 & \texttt{mental\_exec} & Then, after processing, answer[2] is set to 0. Then, second query: l=4-1=3, r=5-1=4. Wait, input is 4 5 5. So l=4-1=3, r=5-1=4. Processing elements $\geq$ 3 and $\leq$ 4 in left [2,3,4]. \\ \hline
204 & \texttt{code\_analysis} & lower\_bound(3) is index 2 (element 3), upper\_bound(4) is index 4 \\ \hline
205 & \texttt{code\_analysis} & (since 4 is in the array, upper\_bound finds first element $>4$, which is index 4. So elements from 2 to 3-1=3-1=3 (wait, the code uses idx\_end. Let me recheck. \\ \hline
206 & \texttt{code\_analysis} & Wait, the code uses: call upper\_bound(left, r, idx\_end) So for the second query, r is 4. \\ \hline
207 & \texttt{mental\_exec} & The left array after first query is [2,3,4]. So upper\_bound(4) will find the first element $>4$. Since all elements are $\leq$ 4, upper\_bound returns size(left)+1=4. \\ \hline
208 & \texttt{mental\_exec} & In the second query, win\_input is 5. So win\_0based is 5-1=4. So answer(4+1=5) is set to 0. So after processing the second query: [Step-through of final answer array logic]. \\ \hline
209 & \texttt{test\_gen} & Testing the code would be necessary. \\ \hline
210 & \texttt{empty} & Now, this code is quite involved. \\ \hline
211 & \texttt{lang\_mapping} & The translation from Java to Fortran required handling the TreeSet with a sorted array and binary search, which increases the code's complexity. But given the constraints, this is the best approach possible. \\ \hline
212 & \texttt{mental\_exec} & Now, let's test this code with a small example. For example, the sample input: 3 1; 1 3 2. \\ \hline
213 & \texttt{high\_level\_plan} & So, the final Fortran code should look like this. \\ \bottomrule
\caption{Complete Reasoning Trace of a CodeLingua L1 Example with Segmented Labels} \\
\end{longtable}

\end{document}